\documentclass[10pt,twocolumn,letterpaper]{article}

\makeatletter
\@namedef{ver@everyshi.sty}{}
\makeatother

\newif\ifarxiv
\arxivtrue % arxiv version
\ifarxiv
  \usepackage[pagenumbers]{cvpr}
  \newcommand{\ARXIVversion}[2]{#1}
\else
  \usepackage{cvpr}
  \newcommand{\ARXIVversion}[2]{#2}
\fi

\usepackage{amssymb}
\usepackage{multirow}
\usepackage{pifont}
\usepackage{xcolor}
\usepackage{xcolor}
\usepackage{tcolorbox}
\usepackage{listings}
\newcommand{\cmark}{\ding{51}}%
\newcommand{\xmark}{\ding{55}}%
\newcommand{\gray}[1]{{\color{gray}#1}}

\usepackage{graphicx}
\usepackage{tikz}
\usepackage{amsmath}
\usepackage{amssymb}
\usepackage{booktabs}
\usepackage{soul}
\usepackage{balance}
\usepackage{flushend}
\usepackage[accsupp]{axessibility}

\usepackage{fancyvrb}
\usepackage{fvextra}

\newcommand{\paragraphcustom}[1]{\vspace{4pt}\noindent\textbf{#1}}
\newcommand{\paragraphcustomWOvspace}[1]{\noindent\textbf{#1}}
\newcommand{\added}[1]{#1}

\definecolor{cvprblue}{rgb}{0.21,0.49,0.74}
\usepackage[pagebackref,breaklinks,colorlinks,allcolors=cvprblue]{hyperref}

\def\paperID{103} % *** Enter the Paper ID here
\def\confName{CVPR}
\def\confYear{2025}

\title{ShowHowTo: Generating Scene-Conditioned Step-by-Step Visual Instructions}

\author{
	Tom\'{a}\v{s} Sou\v{c}ek\textsuperscript{1}
	\quad%\quad\quad
        Prajwal Gatti\textsuperscript{2}
        \quad%\quad\quad
	Michael Wray\textsuperscript{2}
	\quad
	Ivan Laptev\textsuperscript{3}
	\quad%\quad\quad
        Dima Damen\textsuperscript{2}
	\quad%\quad\quad
	Josef Sivic\textsuperscript{1}
	\\
	\small{$^1$CIIRC CTU \quad $^2$University of Bristol \quad $^3$MBZUAI}
	\\
	\small{\texttt{tomas.soucek@cvut.cz}}
	\\
	\small{\url{https://soczech.github.io/showhowto/}}
}

\begin{document}

\twocolumn[{
\maketitle
\vspace{-0.3cm}
\centering
\includegraphics[width=1\textwidth]{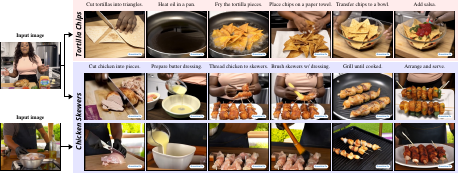}

\captionof{figure}{Given an input image (left) and ordered step-by-step textual instructions for a task (top), \textcolor{cvprblue}{ShowHowTo} generates an image sequence of visual instructions. Rows 1 and 2 demonstrate the generation of visual instructions for two recipes starting from the same input image. Rows 2 and 3 show the generation of visual instructions for the same recipe but conditioned on different input images. \textcolor{cvprblue}{ShowHowTo} generates scene-consistent (\eg, consistency in the person and cutting board) and temporally consistent image sequences (\eg, the bowl of tortilla chips or plate of chicken skewers) that faithfully capture the instructions (\eg, cutting, frying, brushing, adding \etc).}
\label{fig:teaser}
\vspace{0.6cm}
}]
\footnotetext[1]{Czech Institute of Informatics, Robotics and Cybernetics at the Czech Technical University in Prague.}
\footnotetext[2]{School of Computer Science, Machine Learning and Computer Vision (MaVi) Research Group, University of Bristol, UK.}
\footnotetext[3]{Mohamed bin Zayed University of Artificial Intelligence.}

\begin{abstract}
The goal of this work is to generate step-by-step visual instructions in the form of a sequence of images, given an input image that provides the scene context and the sequence of textual instructions. This is a challenging problem as it requires generating multi-step image sequences to achieve a complex goal while being grounded in a specific environment. Part of the challenge stems from the lack of large-scale training data for this problem. The contribution of this work is thus three-fold. 
First, we introduce an automatic approach for collecting large step-by-step visual instruction training data from instructional videos. We apply this approach to one million videos and create a large-scale, high-quality dataset of 0.6M sequences of image-text pairs. Second, we develop and train ShowHowTo, a video diffusion model capable of generating step-by-step visual instructions consistent with the provided input image. Third, we evaluate the generated image sequences across three dimensions of accuracy (step, scene, and task) and show our model achieves state-of-the-art results on all of them. Our code, dataset, and trained models are publicly available.
\end{abstract}

\vspace{-0.2cm}
\section{Introduction}
With the immense success of large vision-language models and the rise of wearable devices, %such as smart glasses and headsets, 
we rapidly approach the era of personalized visual assistants.
This technology promises to help us in a variety of everyday tasks and numerous scenarios, such as preparing a Michelin-star dish, taking care of plants, or fixing a bicycle. %or finding your lost keys. 
Unlike generic instructional videos, visual assistants will provide guidance and feedback for specific environments and task variations. 

Besides being useful for people, automated visual guidance has been recently explored and shown to be beneficial in robotics. For example, \cite{black2023zero,kang2024incorporating,nair2018visual,yu2023scaling}
generate images of intermediate goals and use them as guidance for manipulation policies. Other recent methods~\cite{bharadhwaj2024gen2act,du2024learning,liang2024dreamitate,xu2024flow} derive robotics policies from videos specifically generated for target tasks and environments. 

When comparing the ability to generate textual step-by-step instructions vs. generating visual instructions, one can see a sharp contrast.
State-of-the-art LLMs can reliably provide personalized step-by-step text-only instructions. 
However, translating such instructions into images and videos still presents considerable challenges.
This is because current video generation models, despite their impressive progress over the past years, focus on producing relatively short clips~\cite{sora2024,polyak2024movie,runwayml2024gen3,yang2024cogvideox} whereas image generation models only produce one image at a time.

Recent attempts to generate visual instructions either synthesize a single step~\cite{krojer2024aurora,lai2023lego,soucek2024genhowto} or are not contextualized to the user's specific environment~\cite{menon2024generating,phung2024coherent,bordalo2024generating}.
In other words, such methods may generate plausible images for each step, however, such images will represent arbitrary settings and can feature tools or ingredients unavailable to the user.
In robotics, temporally inconsistent guidance may imply physically implausible demonstrations, resulting in unsuccessful learning of policies. To address this issue, we focus on generating step-by-step visual instructions \textit{conditioned on an input image} from the user, which we assume showcases their starting position---the environment, ingredients, tools, \etc, as illustrated in Figure~\ref{fig:teaser}.

This paper makes the following contributions. (1) We introduce the problem of generating a sequence of visual instructions conditioned on an input image. (2) We introduce a fully automatic approach to collect step-by-step visual instruction training data from in-the-wild instructional videos, creating a large-scale, high-quality dataset of 0.6M step-by-step instruction sequences of 4.5M image-text pairs. (3) We train a video diffusion model capable of generating sparse step-by-step visual instructions consistent with the input image. (4) We evaluate our generated sequences across three aspects (step, scene, and task) and show our model achieves state-of-the-art results on all of them.

\section{Related Work}
\begin{figure*}
    \centering
    \includegraphics[width=\linewidth]{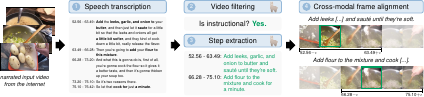}\vspace*{-6pt}
    \caption{\textbf{Our automatic approach for creating the ShowHowTo dataset}---a large-scale instructional dataset consisting of step-by-step instruction sequences of image-text pairs to perform diverse HowTo tasks. 
    Examples of step-by-step textual instructions and the corresponding frames are highlighted in green. 
    }
   \vspace*{-6pt}
    \label{fig:dataset-collection}
\end{figure*}

\paragraphcustomWOvspace{Datasets of visual instructions.}
The scale and quality of the training data play a key role in visual instruction generation. Many available datasets combine instructional or egocentric videos and manual temporal annotation of individual steps~\cite{song2024ego4d, tang2019coin, zhukov2019cross, afouras2024ht} or use professional illustrations~\cite{yang2021visual}. Yet the requirement of manual annotations makes these sources of data hard to scale to novel tasks and environments. To alleviate the need for manual annotations, self-supervised methods have been developed to automatically obtain key steps from in-the-wild videos~\cite{yan2023unloc, mavroudi2023learning, xue2024learning, dvornik2023stepformer, soucek2024multitask}. These key steps can then be used for visual instruction generation~\cite{soucek2024genhowto}. Recently, the improved capabilities of large language models~\cite{dubey2024llama,achiam2023gpt,chowdhery2023palm,NEURIPS2022_8bb0d291,claude3} allowed for solely using video narrations to produce temporal captions, key steps, and instructions~\cite{li2024multi,shvetsova2025howtocaption,lai2023lego}. We build on these works to automatically obtain key steps from videos; however, in contrast to the related works, our dataset is constructed completely automatically, is composed of individual instruction frames instead of temporal intervals, and contains significantly fewer errors. Additionally, we focus on the entire domain of instructional videos, not only cooking.

\paragraphcustom{Conditional video generation.}
Recently, diffusion models~\cite{ho2020denoising} have seen a surge in popularity for generative tasks, including video generation~\cite{yu2023video,mei2023vidm}.
Initial works extended a U-Net model using space-time factorization for the generation of videos in pixel space~\cite{ho2022video,ho2022imagen,hu2022make}.
With the large cost of generating video, others~\cite{chen2023videocrafter1,zhou2022magicvideo,he2022latent,blattmann2023align,ge2023preserve,yu2023video,mei2023vidm} instead use an auto-encoder to model videos within the latent space, reducing significantly the number of parameters and memory requirements.
Video generation models are often conditioned on textual prompts~\cite{girdhar2023emu,chen2024videocrafter2,ho2022video,ho2022imagen,zhang2024show,ge2023preserve}, images that act as initial frames~\cite{yang2023diffusion,zhang2023i2vgen}, or both to generate a sequence of frames~\cite{blattmann2023stable,bar2024lumiere,gupta2023photorealistic,kondratyuk2023videopoet,hu2022make,zhou2022magicvideo,blattmann2023align,chen2023seine,xing2024dynamicrafter}.
It has also been shown that for temporal consistency, a combination of both image conditioning and textual conditioning is important~\cite{wang2024videocomposer,xing2024dynamicrafter}.
However, these methods focus on relatively short video clips and are not able to generate long multi-step sequences of fine-grained instructions that take minutes to execute.

\paragraphcustom{Generating step-by-step instructions.} Both step-by-step visual instructions and continuous videos consist of sequences of frames, yet the instruction sequences differ from the videos significantly. While videos often contain only small pixel-level frame-to-frame changes~\cite{yang2024learning}, visual instruction sequences often contain large semantic (\eg, \textit{raw} $\rightarrow$ \textit{cooked}) and viewpoint (\eg, \textit{inside} $\rightarrow$ \textit{outside}) changes from one key frame to another~\cite{menon2024generating}. Obtaining sufficient training data for visual instructions presents a significant challenge. 
Therefore, Phung \etal \cite{phung2024coherent} generate step-by-step visual instructions using a pretrained text-conditioned image diffusion model with shared attention across steps to ensure consistency in the generated image sequences, while Menon \etal \cite{menon2024generating} use illustrations drawn by artists from WikiHow~\cite{yang2021visual} as the training data for text-to-image-sequence generation. Other works using image-conditioned models can generate step-by-step sequences by iterative generation~\cite{bordalo2024generating,soucek2024genhowto,lai2023lego,krojer2024aurora}. In contrast, our method generates step-by-step visual instructions all at once, attending across steps to generate the full image sequence, including the input, which results in superior quality and consistency.

\section{Building Large-Scale ShowHowTo Dataset}
\label{sec:dataset-collection}

Learning to generate visual instructions requires a large-scale dataset that captures the rich diversity of real-world tasks and their step-by-step execution. However, manually creating such a dataset is prohibitively expensive and time-consuming, limiting the dataset's scale and coverage. We address this challenge by introducing an automated approach that leverages the natural alignment between narrations and visual demonstrations in instructional videos from the web to mine high-quality sequences of image-text pairs. 

Using our proposed approach, we construct a large-scale dataset containing over half a million instruction sequences of image-text pairs spanning 25,026 diverse HowTo tasks. These sequences cover diverse domains including cooking (\eg, \textit{make strawberry crumb bars}, \textit{prepare an avocado margarita}), home improvement (\eg, \textit{stain a cabinet}, \textit{create a tire garden}), assembly (\eg, \textit{set up a 10$\times$10 tent}, \textit{tie a ring sling}), DIY crafts (\eg, \textit{make a bracelet}, \textit{make a fairy glow jar}) and many more. We note that our data collection approach does not require \textit{any} manual annotation, which is an important aspect to enable scaling.

\subsection{Automatic Dataset Collection}

Our approach takes as input a narrated instructional video for a specific task. First, it extracts a sequence of key steps in the form of concise, free-form textual instructions from the video's narration. Then, it associates each step with the corresponding keyframe in the video.
The output is an ordered sequence of image-text pairs. %is automatically extracted from each narrated video. % that matches the task.
This is a very challenging task due to the high level of noise, the possible misalignment of the narration and the visual content as well as the sheer variety of visual appearance and the spoken natural language in the input internet videos. 

To tackle these challenges, we design a four-stage approach, illustrated in Figure \ref{fig:dataset-collection}:  
(1) The narration of the input internet instructional video is transcribed into sentences with corresponding timestamps. (2) The transcribed narration is verified to be instructional and removed if not. (3) The key instruction steps are extracted from the transcript along with their approximate temporal bounds. (4) A representative frame for each instruction step is selected through cross-modal alignment.
By applying this approach to videos from HowTo100M~\cite{miech2019howto100m}, we obtain 578K high-quality sequences of image-text pairs with approximately eight steps per video on average.

Formally, we define our dataset as a collection of instruction sequences of image-text pairs. Each sequence $\{(I_i, \tau_i)\}_{i=0}^{n}$ represents an ordered set of steps required to accomplish a specific task~$\mathcal{T}$. It consists of pairs of images $I_i$ and the corresponding natural language instructions $\tau_i$, with $n$ denoting the number of steps in the sequence. Next, we describe the four stages in detail.

\paragraphcustom{Speech transcription.} Accurate transcription (ASR) of spoken narrations is a key strength of our approach, as these transcripts capture the instructor's step-by-step guidance that we use to align with the video.
HowTo100M provides 1.2 million web instructional videos, but we forego the original transcripts generated using the YouTube API due to noise~\cite{han2022temporal,li2024multi}.
Instead, we use WhisperX~\cite{bain2022whisperx}, a state-of-the-art speech recognition model, to obtain high-quality transcriptions with accurate timestamps from videos.
We provide comparisons of the original transcripts and the improved ones in \ARXIVversion{Figure~\ref{supmatfig:asr_comparison}}{the supplementary material~\cite{soucek25showhowto}}.

\paragraphcustom{Filtering of irrelevant videos.}
We find that many How\-To\-100M videos are non-instructional, containing product reviews, vlogs, movie clips, \etc. This noise may stem from the original data collection process, which relied on keyword-based web crawling and is susceptible to false positives due to inaccurate metadata.
We leverage video transcripts as a strong signal for identifying instructional content and use a recent LLM (Llama 3.1~\cite{dubey2024llama}) to filter the videos.
We verify the reliability of this process through the evaluation on a labeled subset. Detailed analysis, qualitative results, and the prompts used for querying the LLM are provided in \ARXIVversion{Appendix~\ref{supmatsec:dscolldetails} and Figure~\ref{supmatfig:prompt-filtering}}{the supplementary material~\cite{soucek25showhowto}}.

\paragraphcustom{Step extraction.} We observe that, in instructional videos, the key steps necessary to achieve a particular task are very often mentioned in the narration, even if they are not well-aligned with what is shown in the frame~\cite{han2022temporal}. Building on this, we prompt an LLM to extract the instructional steps from the narration transcripts in the format of WikiHow step-by-step guides, providing exemplars in the prompt to guide the extraction. Somewhat surprisingly, the LLM not only correctly extracts the key steps from the transcripts, but the model is also able to associate each step with the correct temporal intervals from the transcript, even if the step spans over multiple narrations. See Figure~\ref{fig:dataset-collection} for an example, and \ARXIVversion{Appendix~\ref{supmatsec:dscolldetails} and Figure~\ref{supmatfig:prompt-step-extraction}}{the supplementary material~\cite{soucek25showhowto}} for additional details and the prompt used.

\paragraphcustom{Cross-modal frame alignment.}
For each instructional step, our goal is to identify a single representative frame that best demonstrates the instruction visually. While contrastive models~\cite{radford2021learning,zhai2023sigmoid,fang2023data} can align text instructions with frames, we observe that naive text matching across thousands of video frames leads to noisy results. Therefore, we limit the alignment to the identified instruction step temporal interval, expanded by $\epsilon=15$ seconds to allow for some level of misalignment between the narrations and visual demonstrations~\cite{han2022temporal}. Given these expanded intervals, we compute text-frame similarity scores using DFN-CLIP~\cite{fang2023data} and select the best alignment that satisfies the temporal ordering of the steps. We provide more details about the matching process in \ARXIVversion{Appendix~\ref{supmatsec:dscolldetails}}{the supplementary material~\cite{soucek25showhowto}}.

\subsection{Dataset Statistics} In total, after filtering, the dataset contains 578K unique sequences of image-text pairs, with a total of 4.5M steps, averaging 7.7~($\pm$~2.8) steps per sequence, and 11.4~($\pm$~4.7) words per step. The sequence lengths vary from 1 to 26 steps, with $97.6\%$ of sequences being 2 to 16 steps long. From task information provided by HowTo100M, the dataset contains instructions for 25K HowTo tasks across several categories, such as cooking, home and garden improvement, vehicles, personal care, health, and more. 
We include a comparative table to other datasets and further dataset analysis in \ARXIVversion{the appendix}{the supplementary material~\cite{soucek25showhowto}}.

\section{ShowHowTo Model and Training Procedure}
\begin{figure}
    \centering
    \includegraphics[width=\linewidth]{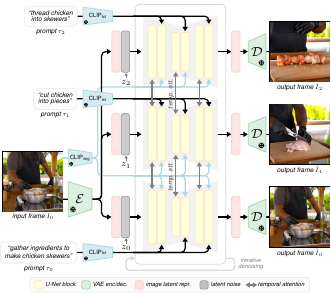}\vspace*{-6pt}
    \caption{\textbf{Model architecture.} Given an input frame $I_0$ (left) and a variable number of text instructions $\tau_i$ describing each step, our diffusion model generates visual instructions $\hat{I}_i$ that correctly follow the prompts $\tau_i$ and are consistent with the input image $I_0$.
    % is conditioned on the image and one text, but ...
    }
    \vspace*{-6pt}
    \label{fig:architecture}
\end{figure}

Given a user-provided image $I_0$, such as a photo of ingredients or tools on a table, our goal is to generate a sequence of images~$\{\hat{I}_i\}_{i=1}^n$ of any length $n$ based on the number of required steps, that guides the user to achieve an intended task~$\mathcal{T}$, such as a cooked chicken tikka masala dish. 
Our goal is to generate images~$\hat{I}_i$ to match the user-provided context, \ie, to be grounded in the user's environment by utilizing the specific objects, tools, and workspace from the input image~$I_0$.
We achieve this goal by training a diffusion model conditioned on the input image $I_0$ along with the step-by-step textual instructions $\{\tau_i\}_{i=0}^n$ that fulfill the intended task~$\mathcal{T}$ in any number of steps $1\leq n \leq 15$. 

We build on recent progress in diffusion models for video generation~\cite{xing2024dynamicrafter}. However, there are the technical challenges of (a) how to inject the multi-step instruction guidance and (b) how to generate variable length sequences. We address these challenges in the next paragraphs.

\paragraphcustom{Architecture.}
Our model, shown in Figure~\ref{fig:architecture}, is based on a latent video diffusion model~\cite{xing2024dynamicrafter} composed of a U-Net encoder and decoder, each with interleaving spatial and temporal attention layers. The input image $I_0$ is projected into the latent space via the VAE encoder $\mathcal{E}$, and is concatenated to each frame of the random noise $z_i$ to form the model's input.
The U-Net progressively denoises the input latent sequence while attending to all images in the sequence to ensure the generated images are temporally consistent and aligned with the input image.
For better conditioning on the input image, the U-Net also contains cross-attention layers that attend to a feature representation of the input image directly.
To guide the generation process to the desired visual instruction sequence, each frame~$i$ in the sequence attends to its prompt $\tau_i$ via cross-attention layers of the U-Net. 
In the ablations, we show that separate text conditioning for each frame in the sequence is instrumental for generating high-quality step-by-step visual instructions.

\paragraphcustom{Training.}
We initialize the model from the pretrained checkpoint~\cite{xing2024dynamicrafter} trained on WebVid10M~\cite{bain2021frozen} for image animation and fine-tune the entire U-Net weights on our dataset. In contrast to training on videos, where the output video is commonly of a fixed length (\eg, 16 frames in the case of \cite{xing2024dynamicrafter}), step-by-step instructions have a variable sequence length. 
To ensure our model can generate variable-length sequences, we vary the sequence length during training by sampling from our sequences. For efficient computation, the length is varied over different batches but is kept the same across all samples in a single batch.
If the dataset sequence is longer than the desired length $k$, we randomly sample the starting frame and use the next consecutive $k$ frames as the model's target. We verify and further discuss these choices in Section~\ref{subsec:ablations} and provide the implementation details in \ARXIVversion{Appendix~\ref{supmatsec:impldetails}}{the supplementary material~\cite{soucek25showhowto}}.

\section{Experiments}
We first introduce our evaluation setup in Section~\ref{subsec:eval-details}. In Section~\ref{subsec:comparison}, we compare our method to current state-of-the-art quantitatively on the test set and through a user study. Section~\ref{subsec:ablations} analyzes key designs of our method through ablation studies. Finally, Section~\ref{subsec:qualitative}, showcases qualitative results of our method. For implementation details and additional results, see \ARXIVversion{the appendix}{the supplementary material~\cite{soucek25showhowto}}.
 
\subsection{Evaluation details}
\label{subsec:eval-details}

\paragraphcustomWOvspace{Dataset.}
We construct train and test splits from our dataset of 578K samples. Our test set comprises 3,964 sequences from 200 tasks covering the distribution over task categories in the full dataset.
To ensure sample quality, we prioritize samples with high DFN-CLIP alignment scores as measured in our dataset creation pipeline (Section~\ref{sec:dataset-collection}). \added{For zero-shot evaluation of our method, we also use a random subset of 1442 non-illustrated instructional sequences\footnote{See the project website for the list of selected sequences.} from the WikiHow-VGSI dataset~\cite{yang2021visual} as an additional test set.}

\paragraphcustom{Evaluation metrics.}
We evaluate our model by measuring the correctness and consistency of the generated visual instructions using similar metrics as Menon \etal~\cite{menon2024generating}. 
We describe the used metrics next with full details available in \ARXIVversion{Appendix~\ref{supmatsec:evaldetails}}{the supplementary material~\cite{soucek25showhowto}}.
\textbf{(1) Step Faithfulness}~\cite{menon2024generating} measures whether each generated image $\hat{I}_i$ correctly depicts its corresponding text instruction $\tau_i$. It is computed as the zero-shot accuracy of the DFN-CLIP model where the generated image $\hat{I}_i$ is classified into classes $\{\tau_i\}_{i=0}^{n}$ of all text instructions of the sequence.
\textbf{(2) Scene Consistency} measures whether the generated image $\hat{I}_i$ consistently captures the scene from the input image $I_0$ (\eg, the same utensils are used on the same kitchen countertop as in the input image). Intuitively, a generated image $\hat{I}_i$ is considered scene-consistent if it visually matches any frame from its source video $\{I_i\}_{i=1}^{n}$ (excluding the input image to avoid trivial copy solution). Therefore, for each generated image $\hat{I}_i$, the most similar image according to the DINOv2~\cite{oquab2023dinov2} score is retrieved from the test set images. The metric then measures if the retrieved image is from the same video as the input.
\textbf{(3) Task Faithfulness} measures how well the generated sequence $\{\hat{I}_i\}_{i=1}^{n}$ represents its intended task. It is measured as the zero-shot accuracy of the DFN-CLIP model where the generated sequence's averaged feature vector is classified into all 200 test set tasks. In contrast to Menon \etal~\cite{menon2024generating}, the generated sequences are evaluated holistically rather than per-step, as often steps are not unique to a task (\eg, ``knead dough'' step appears in both ``Make sourdough bread'' and ``Make pizza'' tasks), and the classification is done into all test set tasks rather than a small random subset, providing a more robust evaluation.
%
% Lastly, we report \textbf{Overall} score as the average of the three metrics.

\subsection{Comparison with the State-of-the-Art}
\label{subsec:comparison}

\begin{table}[t]
\centering
\scriptsize
\begin{tabular}{c@{\hskip 0.1cm}lc@{~~}c@{~~}cc@{~~}c}
\toprule
& & \multicolumn{3}{c}{ShowHowTo} & \multicolumn{2}{c}{WikiHow~\cite{yang2021visual}} \\
\cmidrule(lr){3-5} \cmidrule(lr){6-7}
&\multirow{2}{*}{Method}  & {Step} &  {Scene} & {Task} & {Step} &  {Scene} \\
&& {Faithf.} &  {Consist.} & {Faithf.} & {Faithf.} &  {Consist.}  \\
\midrule
\textbf{(a)} & InstructPix2Pix~\cite{brooks2023instructpix2pix} & 0.25 & 0.17 & 0.25 & 0.32 & 0.12 \\
\textbf{(b)} & AURORA~\cite{krojer2024aurora} & 0.25 & 0.33 & 0.24 & 0.33 & \textbf{0.15} \\
\textbf{(c)} & GenHowTo~\cite{soucek2024genhowto} & 0.49 & 0.13 & 0.27 & 0.60 & 0.06 \\
\textbf{(d)} & Phung \etal~\cite{phung2024coherent} & 0.36 & 0.03 & 0.38 & 0.46 & 0.04 \\
\textbf{(e)} & StackedDiffusion~\cite{menon2024generating} & 0.43 & 0.02 & \textbf{0.42} & 0.57 & 0.07 \\
\textbf{(f)} & \textbf{ShowHowTo} & \textbf{0.52} & \textbf{0.34} & \textbf{0.42} & \textbf{0.72} & {0.12} \\
\midrule
\textbf{(g)} & \gray{\textit{Random}} & \gray{0.19} & \gray{0.00} & \gray{0.01} & \gray{0.26} & \gray{0.00} \\
\textbf{(h)} & \gray{Stable Diffusion~\cite{rombach2022high}\textsuperscript{$\dagger$}} & \gray{0.70} & \gray{0.03} & \gray{0.44} & \gray{0.84} & \gray{0.03} \\
\textbf{(i)} & \gray{\textit{Copy of the input image}} & \gray{0.19} & \gray{{0.62}} & \gray{0.39} & \gray{0.26} & \gray{{0.26}} \\
\textbf{(j)} & \gray{\textit{Source sequences}} & \gray{{0.50}} & \gray{1.00} & \gray{{0.56}} & \gray{{0.60}} & \gray{1.00} \\
\bottomrule
\multicolumn{7}{c}{\textsuperscript{$\dagger$}~{Generation not conditioned on the input image.}}\\
\end{tabular}\vspace*{-6pt}
\caption{\textbf{Comparison with state-of-the-art on the ShowHowTo and the WikiHow datasets}.
Out of all the visual instruction generation methods, our method best follows the input prompts while being consistent with the input image.}
\vspace*{-6pt}
\label{tab:main-results}
\end{table}

\paragraphcustomWOvspace{Compared methods.} We compare ShowHowTo to state-of-the-art methods for visual instruction generation as well as various baselines. For image-to-image methods \textbf{(a-c)}, we generate the visual instructions sequence by iteratively using the last generated image as the input for the next step generation \added{to achieve temporal consistency}. \textbf{(a) InstructPix2Pix}~\cite{brooks2023instructpix2pix} is trained to manipulate input images according to a text prompt by training on synthetic paired image data. In contrast, \textbf{(b) AURORA}~\cite{krojer2024aurora} is trained on a manually curated dataset of image pairs from videos, while \textbf{(c) GenHowTo}~\cite{soucek2024genhowto} extracts the image pairs for training from instructional videos automatically.

Methods that generate image sequences \textbf{(d-e)} do not accept an input image, therefore, we apply the common input masking approach, where the first denoised frame of the sequence is replaced by the noised ground truth frame in each step of the generation. The \textbf{(d)~Phung \etal}~\cite{phung2024coherent} method generates consistent sequences of visual instructions by attending to all frames in the sequence in the spatial attention layers. As it is a training-free method, we reimplement it and use it with the Stable Diffusion backbone~\cite{rombach2022high}. \textbf{(e) StackedDiffusion}~\cite{menon2024generating} is trained on WikiHow illustrated image sequences. It generates the image sequence as a single tiled image. \added{Similarly to the related work, we evaluate all methods in zero-shot setup without finetuning.}

Lastly, we show \textbf{(g) Random} lower bound and various naive baselines \textbf{(h-i)}. \textbf{(h) Stable Diffusion}~\cite{rombach2022high} is a text-conditioned generative method with no input image conditioning that generates each image independently, the \textbf{(i)~Copy} baseline uses the input image as the output for any prompt. As an upper limit, \textbf{(j) Source sequences} uses the original dataset frames corresponding to the text prompts.

\paragraphcustom{Quantitative results.} We show the results of different methods on our ShowHowTo test set \added{as well as on a subset of WikiHow sequences~\cite{yang2021visual}} in Table~\ref{tab:main-results}.
The methods trained to perform localized edits \textbf{(a-b)} generate outputs fairly consistent with the input image (see the Scene Consistency metric), yet they fail to properly capture the instructions described by the text prompts (evidenced by the Step Faithfulness metric).
On the other hand, methods for generating sequences of visual instructions \textbf{(d-e)} model the instructions well according to the input text prompts, but they perform poorly in scene consistency. In contrast, our method \textbf{(f)} generates visual instructions that are consistent with the input scene and correctly capture the action specified by the prompt. Our method even generates images that are more faithful to the input textual instructions than the dataset sequences \textbf{(j)} (see the Step Faithfulness metric). 
\added{There are two reasons for this: objects in real images can appear small or occluded, impacting CLIP matching, and sometimes steps do not appear visually in the video.}

\added{Additionally, in \ARXIVversion{Appendix~\ref{supmatsec:addquantres}}{the supplementary material~\cite{soucek25showhowto}}, we test our method in a zero-shot setup on the GenHowTo benchmark~\cite{soucek2024genhowto} and report additional quantitative metrics on the ShowHowTo dataset.}

\begin{figure}
    \centering
    \includegraphics[width=\linewidth]{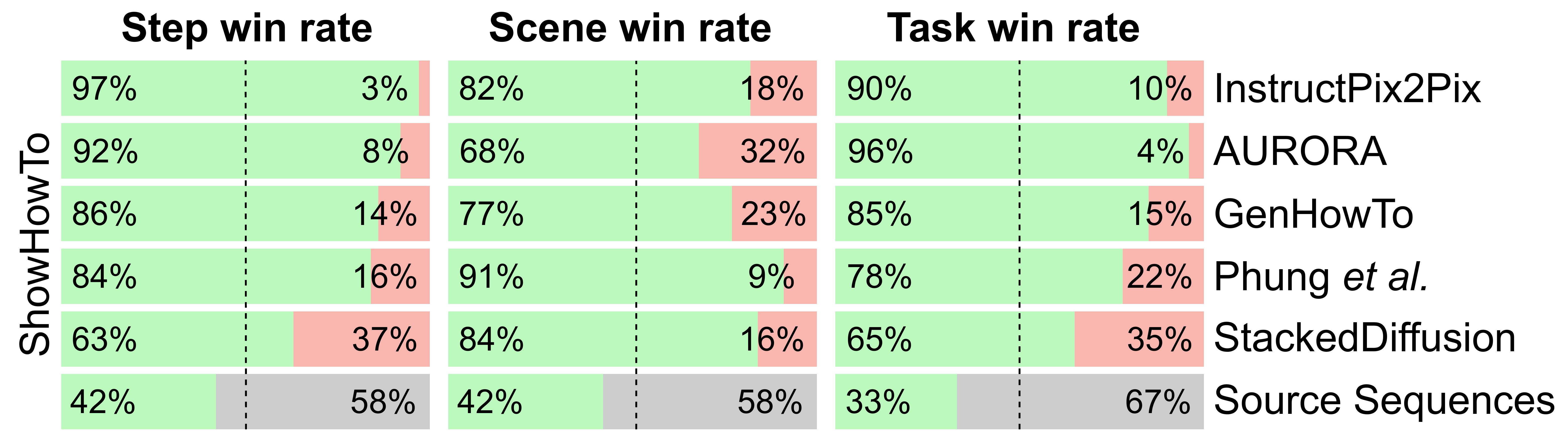}\vspace*{-6pt}
    \caption{\textbf{User study results.} Win rates of the ShowHowTo method against baselines from pairwise forced decision user evaluations, divided into step, scene, and task. Values larger than 50\% indicate ShowHowTo is preferred over the other methods (right).
    }
    \vspace*{-6pt}
    \label{fig:user-study}
\end{figure}

\paragraphcustom{User study.}
We present a user study with 9 participants evaluating sequences from 100 randomly sampled tasks from our test set. Each participant compared 50 ShowHowTo sequences with baselines using three criteria: (1) Step Faithfulness (\textit{Which sequence better follows the steps?}), (2) Scene Consistency (\textit{Which sequence is more likely to come from the same video?}), and (3) Task Faithfulness (\textit{Which sequence accurately depicts the instructions for the task of \texttt{[task]}?}). As shown in Figure~\ref{fig:user-study}, ShowHowTo outperforms all baselines. Notably, users preferred our generations over sequences from source videos in 42\% of cases for both step and scene metrics, which may be attributed to instructional videos not showing good views of steps at times and the high quality of our generation. Lower task faithfulness scores against the source sequences suggest room for improvement in future methods. More details are in \ARXIVversion{Appendix~\ref{supmatsec:evaldetails}}{the supplementary material~\cite{soucek25showhowto}}.

\subsection{Ablations}
\label{subsec:ablations}
We evaluate the key design decisions of our proposed method, \ie, the model conditioning, training data, and variable sequence length training, in the next paragraphs. \added{Furthermore, additional performance analysis of the trained model is available \ARXIVversion{Appendix~\ref{supmatsec:addquantres}}{the supplementary material~\cite{soucek25showhowto}}.}

\begin{table}[t]
\centering
\scriptsize
\begin{tabular}{lc@{\hskip 0.15cm}c@{\hskip 0.15cm}c@{\hskip 0.15cm}c}
\toprule
\multirow{2}{*}{Text conditioning type} & Step & Scene & Task & \multirow{2}{*}{Average} \\
& {Faithf.} &  {Consist.} & {Faithf.} &  \\
\midrule
1 prompt (concatenated step prompts) & 0.21 & 0.29 & 0.38 & 0.29 \\
1 prompt (summarized step prompts) & 0.20 & 0.30 & 0.40 & 0.30 \\
1 prompt per step ($\tau_0=$ \texttt{`an image'})  & {0.51} & {0.30} & \textbf{0.42} & {0.41} \\
1 prompt per step (\textbf{ShowHowTo})  & \textbf{0.52} & \textbf{0.34} & \textbf{0.42} & \textbf{0.43} \\
\bottomrule
\end{tabular}
\ARXIVversion{}{\vspace{-0.2cm}}\vspace*{-2pt}
\caption{\textbf{Ablation of step conditioning}. The per-frame conditioning of ShowHowTo is instrumental in generating visual instructions faithful to the textual instructions.}
\label{tab:ablation-textconditioning}
\end{table}

\paragraphcustom{Text model conditioning.}
We evaluate how different types of text conditioning affect model performance. For video models, it is common to provide a single text prompt for conditioning. However, visual instructions vary substantially from one another, possibly requiring different conditioning. We construct a single prompt for each sequence by concatenating all step prompts and by summarizing the step prompts using an LLM~\cite{dubey2024llama}. In Table~\ref{tab:ablation-textconditioning}, we show that our choice of separate prompt per step significantly outperforms both of the single prompt variants. Additionally, we demonstrate that 
using the free-form step description for $\tau_0$ outperforms the fixed prompt \texttt{`an image'}.

\begin{table}[t]
\centering
\scriptsize
\begin{tabular}{lcccc}
\toprule
\multirow{2}{*}{Model training data}   & Step & Scene & Task & \multirow{2}{*}{Average}  \\
& {Faithf.} &  {Consist.} & {Faithf.} &  \\
\midrule
% ShowHowTo (10\% of the data) \\
% \midrule
WikiHow-VGSI~\cite{yang2021visual} & \textbf{0.55} & 0.12 & 0.30 & 0.32 \\
HowToStep~\cite{li2024multi} & 0.39 & 0.33 & 0.29 & 0.34 \\
ShowHowTo (\textit{food} videos only) & 0.51 & 0.32 & 0.37 & 0.40 \\
\textbf{ShowHowTo} & {0.52} & \textbf{0.34} & \textbf{0.42} & \textbf{0.43} \\
\bottomrule
\end{tabular}
\ARXIVversion{}{\vspace{-0.2cm}}\vspace*{-2pt}
\caption{\textbf{Ablation of the training data} as measured on the ShowHowTo test set. Our training dataset yields significant improvement over the manually curated WikiHow as well as the closely related HowToStep due to the quality of our instructions.}
\vspace*{-6pt}
% \ARXIVversion{}{\vspace{-0.1cm}}
\label{tab:ablation-trainingdata}
\end{table}

\paragraphcustom{Training data.} 
In Table~\ref{tab:ablation-trainingdata}, we analyze the impact of different training datasets by comparing our dataset with two related instructional datasets of similar scale and task coverage: (i) HowToStep~\cite{li2024multi}, which contains automatically extracted video-text sequences from cooking videos, and (ii) WikiHow-VGSI~\cite{yang2021visual}, which consists of manually created image-text sequences from WikiHow articles, where the images primarily consist of digitally drawn illustrations.
To train on the HowToStep dataset, we select the middle frame of each video segment as the visual instruction frame. We observe significantly worse performance caused both by the lack of precise instruction frame information as well as very noisy video segments (\eg, the dataset contains \textit{`Thank you for watching!'} segments which are not instructional). \added{The performance is also worse when compared to the model trained only on the \textit{food}-related ShowHowTo sequences that are extracted from the very same videos as the HowToStep sequences, indicating a superiority of our sequence extraction process.} Training on Wiki\-How-VGSI yields higher Step Faithfulness score, likely due to the quality of manually matched images and prompts. However, the overall performance remains significantly below our approach, as the model primarily learns from illustrated images, resulting in less consistent scene generation. %Lastly, we compare the performance of the ShowHowTo model trained only on cooking task videos---a decision that other works make (\eg, Li \etal~\cite{li2024multi})---with our full training dataset showing our model benefits from scaling the data to the open task domain.

\begin{table}[t]
\centering
\scriptsize
\begin{tabular}{lcccc}
\toprule
\multirow{2}{*}{Training sequence length}  & Step & Scene & Task & \multirow{2}{*}{Average} \\
& {Faithf.} &  {Consist.} & {Faithf.} &  \\
\midrule
$\leq$ 4 steps  & 0.47 & \textbf{0.39} & 0.40 & 0.42 \\
$\leq$ 8 steps (\textbf{ShowHowTo}) & {0.52} & {0.34} & \textbf{0.42} & \textbf{0.43} \\
$\leq$ 8 steps, randomly sampled & 0.51 & 0.32 & 0.41 & 0.41 \\
$=$ 8 steps & 0.56 & 0.26 & \textbf{0.42} & 0.41 \\
$\leq$ 16 steps & \textbf{0.57} & 0.26 & \textbf{0.42} & 0.41 \\
\bottomrule
\end{tabular}
\ARXIVversion{}{\vspace{-0.2cm}}
\caption{\textbf{Ablation of the training sequence length} as measured on the ShowHowTo test set. We compare the performance of our model when trained on different sequence lengths.}
\vspace*{-12pt}
\label{tab:ablation-seqlen}
\end{table}

\paragraphcustom{Variable sequence length.}
Instructions are often of variable length, therefore, one of the key model requirements is to support generating image sequences of different lengths. While attention-based architectures allow for any sequence length, the question is how to train such a model. We test training the model on variable sequences of up to 4 frames, 8 frames, and 16 frames. We also train the model on sequences of length 8 only. In Table~\ref{tab:ablation-seqlen}, we show that training on shorter sequences up to 4 frames results in high Scene Consistency but low Step Correctness, while training on sequences up to 16 frames is the opposite. This can be attributed to the fact that short sequences often keep the same background across the whole sequence, encouraging the model to preserve the background at the expense of the prompt. Longer sequences, on the other hand, have more variation of the background, \eg, as the task moves from the counter to the hob. The model is thus less likely to enforce the background during inference. Lastly, we also show that it is important to train always on consecutive sequences of visual instructions. If a subset of visual instructions from a video is sampled randomly (with the temporal ordering preserved), the scene consistency is decreased (Table~\ref{tab:ablation-seqlen}, row~3).

\begin{figure}
    \centering
    \includegraphics[width=\linewidth]{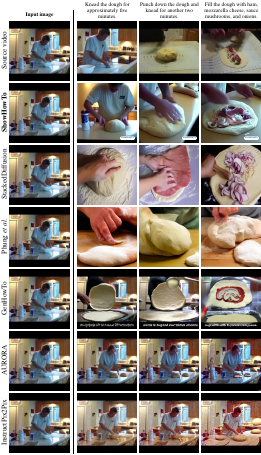}
    \vspace{-0.5cm}
    \caption{\textbf{Qualitative comparison} using the input image (left) and the textual instructions (top) for the task of \textit{making a calzone}. The images from the source video are shown in the first row. Except for ShowHowTo, methods either struggle to preserve the input scene or to produce coherent steps.}
    % \caption{\textbf{Qualitative comparison} using the input image (left) and text instructions (top) for the task of \textit{making a calzone}. First row shows images from the source video. All methods except ShowHowTo either fail to preserve the input scene or produce coherent steps. }
    
    \vspace{-0.5cm}
    \label{fig:qualcomp}
\end{figure}

\subsection{Qualitative Results}
\label{subsec:qualitative}

\begin{figure*}
    \centering
    \includegraphics[width=\linewidth]{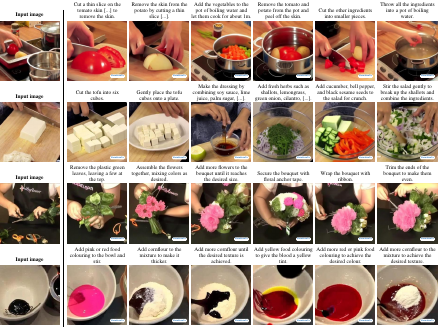}\vspace*{-6pt}
    \caption{\textbf{Qualitative results of our method} for sequences from the test set. Given the input image (left) and the textual instructions (top), ShowHowTo generates step-by-step visual instructions while maintaining objects from the input image (\eg, the cooking pot and the ceramic bowl in rows one and four) as well as among generated images (\eg, glass bowl in the second row). 
    }
    \vspace*{-6pt}
    \label{fig:qualitative-results}
\end{figure*}

Qualitative results in Figure~\ref{fig:teaser}, Figure~\ref{fig:qualitative-results}, and \ARXIVversion{Figures~\ref{supmatfig:qualres1}, \ref{supmatfig:qualres3}, and \ref{supmatfig:qualres2} in the appendix}{additional figures in the supplementary material~\cite{soucek25showhowto}} demonstrate the key strengths of the ShowHowTo model. It consistently preserves the scene as well as various objects, tools, and ingredients from the user-provided input image (\eg, pot and vegetables in Figure~\ref{fig:qualitative-results}, first row). It dynamically adjusts the viewpoint to emphasize key actions. \added{Similarly to vanilla text-to-image and text-to-video models, }our model can also introduce plausible task-relevant objects (\eg, knives or bowls) if these objects are not present in the user-provided input image. Notably, the model effectively adapts human poses to demonstrate various object manipulations (\eg, flower arranging in Figure~\ref{fig:qualitative-results}, third row).

% We also compare ShowHowTo with the related methods qualitatively in Figure~\ref{fig:qualcomp}.
Figure~\ref{fig:qualcomp} shows qualitative comparisons with related methods. Methods for generating instructional sequences (StackedDiffusion~\cite{menon2024generating} and Phung \etal~\cite{phung2024coherent}) fail to preserve the scene from the input image. For example, they generate blue, black, or wooden kitchen countertop instead of the glossy white one from the input image. On the other hand, image-to-image approaches (GenHowTo~\cite{soucek2024genhowto}, AURORA~\cite{krojer2024aurora}, and InstructPix2Pix~\cite{brooks2023instructpix2pix}) perform minimal edits and propagate errors through the output image sequence due to iterative generation (\eg, the persistent floating dough generated by the GenHowTo method). 

\paragraphcustom{Limitations.} 
While our method can generate complex step-by-step visual instructions conditioned on the input image, it inherits the limitations of the models it is based on and introduces new limitations stemming from the novel source of training data. ShowHowTo model can struggle to maintain object states across many frames; for example, it can generate an image with raw meat after it was cooked in previous steps. Though the model often correctly generates common objects from instructional videos, for rare objects, such as electrical components, the model may generate objects in physically impossible configurations. Please see \ARXIVversion{Figure~\ref{supmatfig:failures} in the appendix}{the supplementary material~\cite{soucek25showhowto}} for failure case examples.

\section{Conclusion}

This work explores, for the first time, generating environment-specific visual instructions to accomplish a user-defined task. We introduce a fully automated and scalable pipeline to create a dataset of 578K instructional image-text sequences from online videos, \textit{without requiring any manual supervision}. Using this data, we train the ShowHowTo model to generate contextualized step-by-step visual instructions. Experiments demonstrate superior ability to generate accurate, scene-consistent instructional steps across various HowTo tasks, outperforming existing methods. We believe this work opens new avenues for personalized guidance in assistive technologies and step-by-step goal generation for robot planning.

%\paragraphcustom{Acknowledgements.}
\section*{Acknowledgements}
{\small
We acknowledge VSB – Technical University of Ostrava, IT4Innovations National Supercomputing Center, Czech Republic, for awarding this project access to the LUMI supercomputer, owned by the EuroHPC Joint Undertaking, hosted by CSC (Finland) and the LUMI consortium through the Ministry of Education, Youth and Sports of the Czech Republic through the e-INFRA CZ (grant ID: 90254).
% This work was supported by the Ministry of Education, Youth and Sports of the Czech Republic through the e-INFRA CZ (ID:90140).
% This work was partly supported by the EU Horizon Europe Programme under the project EXA4MIND (No. 101092944) and the Ministry of Education, Youth and Sports of the Czech Republic through the e-INFRA CZ (ID:90140).

Research at the University of Bristol is supported by EPSRC UMPIRE (EP/T004991/1) and EPSRC PG Visual AI (EP/T028572/1). Prajwal Gatti is partially funded by an uncharitable donation from Adobe Research to the University of Bristol. 

This research was co-funded by the European Union (ERC FRONTIER, No. 101097822 and ELIAS No. 101120237) and received the support of the EXA4MIND project, funded by the European Union’s Horizon Europe Research and Innovation Programme, under Grant Agreement N° 101092944. Views and opinions expressed are however those of the author(s) only and do not necessarily reflect those of the European Union or the European Commission. Neither the European Union nor the granting authority can be held responsible for them.
}

{\small
\bibliographystyle{ieeenat_fullname}
\bibliography{main}

\begin{thebibliography}{70}
\providecommand{\natexlab}[1]{#1}
\providecommand{\url}[1]{\texttt{#1}}
\expandafter\ifx\csname urlstyle\endcsname\relax
  \providecommand{\doi}[1]{doi: #1}\else
  \providecommand{\doi}{doi: \begingroup \urlstyle{rm}\Url}\fi

\bibitem[Achiam et~al.(2023)Achiam, Adler, Agarwal, Ahmad, Akkaya, Aleman, Almeida, Altenschmidt, Altman, Anadkat, et~al.]{achiam2023gpt}
Josh Achiam, Steven Adler, Sandhini Agarwal, Lama Ahmad, Ilge Akkaya, Florencia~Leoni Aleman, Diogo Almeida, Janko Altenschmidt, Sam Altman, Shyamal Anadkat, et~al.
\newblock Gpt-4 technical report.
\newblock \emph{arXiv preprint arXiv:2303.08774}, 2023.

\bibitem[Afouras et~al.(2024)Afouras, Mavroudi, Nagarajan, Wang, and Torresani]{afouras2024ht}
Triantafyllos Afouras, Effrosyni Mavroudi, Tushar Nagarajan, Huiyu Wang, and Lorenzo Torresani.
\newblock Ht-step: Aligning instructional articles with how-to videos.
\newblock \emph{NeurIPS}, 2024.

\bibitem[Anthropic(2024)]{claude3}
Anthropic.
\newblock Introducing computer use, a new claude 3.5 sonnet, and claude 3.5 haiku.
\newblock \url{https://www.anthropic.com/news/3-5-models-and-computer-use}, 2024.

\bibitem[Bain et~al.(2021)Bain, Nagrani, Varol, and Zisserman]{bain2021frozen}
Max Bain, Arsha Nagrani, G{\"u}l Varol, and Andrew Zisserman.
\newblock Frozen in time: A joint video and image encoder for end-to-end retrieval.
\newblock In \emph{ICCV}, 2021.

\bibitem[Bain et~al.(2023)Bain, Huh, Han, and Zisserman]{bain2022whisperx}
Max Bain, Jaesung Huh, Tengda Han, and Andrew Zisserman.
\newblock Whisperx: Time-accurate speech transcription of long-form audio.
\newblock \emph{INTERSPEECH}, 2023.

\bibitem[Bar-Tal et~al.(2024)Bar-Tal, Chefer, Tov, Herrmann, Paiss, Zada, Ephrat, Hur, Liu, Raj, et~al.]{bar2024lumiere}
Omer Bar-Tal, Hila Chefer, Omer Tov, Charles Herrmann, Roni Paiss, Shiran Zada, Ariel Ephrat, Junhwa Hur, Guanghui Liu, Amit Raj, et~al.
\newblock Lumiere: A space-time diffusion model for video generation.
\newblock \emph{arXiv preprint arXiv:2401.12945}, 2024.

\bibitem[Bharadhwaj et~al.(2024)Bharadhwaj, Dwibedi, Gupta, Tulsiani, Doersch, Xiao, Shah, Xia, Sadigh, and Kirmani]{bharadhwaj2024gen2act}
Homanga Bharadhwaj, Debidatta Dwibedi, Abhinav Gupta, Shubham Tulsiani, Carl Doersch, Ted Xiao, Dhruv Shah, Fei Xia, Dorsa Sadigh, and Sean Kirmani.
\newblock Gen2act: Human video generation in novel scenarios enables generalizable robot manipulation.
\newblock \emph{arXiv preprint arXiv:2409.16283}, 2024.

\bibitem[Black et~al.(2023)Black, Nakamoto, Atreya, Walke, Finn, Kumar, and Levine]{black2023zero}
Kevin Black, Mitsuhiko Nakamoto, Pranav Atreya, Homer Walke, Chelsea Finn, Aviral Kumar, and Sergey Levine.
\newblock Zero-shot robotic manipulation with pretrained image-editing diffusion models.
\newblock \emph{arXiv preprint arXiv:2310.10639}, 2023.

\bibitem[Blattmann et~al.(2023{\natexlab{a}})Blattmann, Dockhorn, Kulal, Mendelevitch, Kilian, Lorenz, Levi, English, Voleti, Letts, et~al.]{blattmann2023stable}
Andreas Blattmann, Tim Dockhorn, Sumith Kulal, Daniel Mendelevitch, Maciej Kilian, Dominik Lorenz, Yam Levi, Zion English, Vikram Voleti, Adam Letts, et~al.
\newblock Stable video diffusion: Scaling latent video diffusion models to large datasets.
\newblock \emph{arXiv preprint arXiv:2311.15127}, 2023{\natexlab{a}}.

\bibitem[Blattmann et~al.(2023{\natexlab{b}})Blattmann, Rombach, Ling, Dockhorn, Kim, Fidler, and Kreis]{blattmann2023align}
Andreas Blattmann, Robin Rombach, Huan Ling, Tim Dockhorn, Seung~Wook Kim, Sanja Fidler, and Karsten Kreis.
\newblock Align your latents: High-resolution video synthesis with latent diffusion models.
\newblock In \emph{CVPR}, 2023{\natexlab{b}}.

\bibitem[Bordalo et~al.(2024)Bordalo, Ramos, Val{\'e}rio, Gl{\'o}ria-Silva, Bitton, Yarom, Szpektor, and Magalhaes]{bordalo2024generating}
Jo{\~a}o Bordalo, Vasco Ramos, Rodrigo Val{\'e}rio, Diogo Gl{\'o}ria-Silva, Yonatan Bitton, Michal Yarom, Idan Szpektor, and Joao Magalhaes.
\newblock Generating coherent sequences of visual illustrations for real-world manual tasks.
\newblock \emph{arXiv preprint arXiv:2405.10122}, 2024.

\bibitem[Brooks et~al.(2023)Brooks, Holynski, and Efros]{brooks2023instructpix2pix}
Tim Brooks, Aleksander Holynski, and Alexei~A Efros.
\newblock Instructpix2pix: Learning to follow image editing instructions.
\newblock In \emph{CVPR}, 2023.

\bibitem[Brooks et~al.(2024)Brooks, Peebles, Holmes, DePue, Guo, Jing, Schnurr, Taylor, Luhman, Luhman, Ng, Wang, and Ramesh]{sora2024}
Tim Brooks, Bill Peebles, Connor Holmes, Will DePue, Yufei Guo, Li Jing, David Schnurr, Joe Taylor, Troy Luhman, Eric Luhman, Clarence Ng, Ricky Wang, and Aditya Ramesh.
\newblock Video generation models as world simulators.
\newblock 2024.

\bibitem[Chen et~al.(2023{\natexlab{a}})Chen, Xia, He, Zhang, Cun, Yang, Xing, Liu, Chen, Wang, et~al.]{chen2023videocrafter1}
Haoxin Chen, Menghan Xia, Yingqing He, Yong Zhang, Xiaodong Cun, Shaoshu Yang, Jinbo Xing, Yaofang Liu, Qifeng Chen, Xintao Wang, et~al.
\newblock Videocrafter1: Open diffusion models for high-quality video generation.
\newblock \emph{arXiv preprint arXiv:2310.19512}, 2023{\natexlab{a}}.

\bibitem[Chen et~al.(2024)Chen, Zhang, Cun, Xia, Wang, Weng, and Shan]{chen2024videocrafter2}
Haoxin Chen, Yong Zhang, Xiaodong Cun, Menghan Xia, Xintao Wang, Chao Weng, and Ying Shan.
\newblock Videocrafter2: Overcoming data limitations for high-quality video diffusion models.
\newblock In \emph{CVPR}, 2024.

\bibitem[Chen et~al.(2023{\natexlab{b}})Chen, Wang, Zhang, Zhuang, Ma, Yu, Wang, Lin, Qiao, and Liu]{chen2023seine}
Xinyuan Chen, Yaohui Wang, Lingjun Zhang, Shaobin Zhuang, Xin Ma, Jiashuo Yu, Yali Wang, Dahua Lin, Yu Qiao, and Ziwei Liu.
\newblock Seine: Short-to-long video diffusion model for generative transition and prediction.
\newblock In \emph{ICLR}, 2023{\natexlab{b}}.

\bibitem[Chowdhery et~al.(2023)Chowdhery, Narang, Devlin, Bosma, Mishra, Roberts, Barham, Chung, Sutton, Gehrmann, et~al.]{chowdhery2023palm}
Aakanksha Chowdhery, Sharan Narang, Jacob Devlin, Maarten Bosma, Gaurav Mishra, Adam Roberts, Paul Barham, Hyung~Won Chung, Charles Sutton, Sebastian Gehrmann, et~al.
\newblock Palm: Scaling language modeling with pathways.
\newblock \emph{Journal of Machine Learning Research}, 2023.

\bibitem[Du et~al.(2023)Du, Yang, Dai, Dai, Nachum, Tenenbaum, Schuurmans, and Abbeel]{du2024learning}
Yilun Du, Sherry Yang, Bo Dai, Hanjun Dai, Ofir Nachum, Josh Tenenbaum, Dale Schuurmans, and Pieter Abbeel.
\newblock Learning universal policies via text-guided video generation.
\newblock \emph{NeurIPS}, 2023.

\bibitem[Dubey et~al.(2024)Dubey, Jauhri, Pandey, Kadian, Al-Dahle, Letman, Mathur, Schelten, Yang, Fan, et~al.]{dubey2024llama}
Abhimanyu Dubey, Abhinav Jauhri, Abhinav Pandey, Abhishek Kadian, Ahmad Al-Dahle, Aiesha Letman, Akhil Mathur, Alan Schelten, Amy Yang, Angela Fan, et~al.
\newblock The llama 3 herd of models.
\newblock \emph{arXiv preprint arXiv:2407.21783}, 2024.

\bibitem[Dvornik et~al.(2023)Dvornik, Hadji, Zhang, Derpanis, Wildes, and Jepson]{dvornik2023stepformer}
Nikita Dvornik, Isma Hadji, Ran Zhang, Konstantinos~G Derpanis, Richard~P Wildes, and Allan~D Jepson.
\newblock Stepformer: Self-supervised step discovery and localization in instructional videos.
\newblock In \emph{CVPR}, 2023.

\bibitem[Fang et~al.(2023)Fang, Jose, Jain, Schmidt, Toshev, and Shankar]{fang2023data}
Alex Fang, Albin~Madappally Jose, Amit Jain, Ludwig Schmidt, Alexander Toshev, and Vaishaal Shankar.
\newblock Data filtering networks.
\newblock \emph{arXiv preprint arXiv:2309.17425}, 2023.

\bibitem[Ge et~al.(2023)Ge, Nah, Liu, Poon, Tao, Catanzaro, Jacobs, Huang, Liu, and Balaji]{ge2023preserve}
Songwei Ge, Seungjun Nah, Guilin Liu, Tyler Poon, Andrew Tao, Bryan Catanzaro, David Jacobs, Jia-Bin Huang, Ming-Yu Liu, and Yogesh Balaji.
\newblock Preserve your own correlation: A noise prior for video diffusion models.
\newblock In \emph{ICCV}, 2023.

\bibitem[Girdhar et~al.(2023)Girdhar, Singh, Brown, Duval, Azadi, Rambhatla, Shah, Yin, Parikh, and Misra]{girdhar2023emu}
Rohit Girdhar, Mannat Singh, Andrew Brown, Quentin Duval, Samaneh Azadi, Sai~Saketh Rambhatla, Akbar Shah, Xi Yin, Devi Parikh, and Ishan Misra.
\newblock Emu video: Factorizing text-to-video generation by explicit image conditioning.
\newblock \emph{arXiv preprint arXiv:2311.10709}, 2023.

\bibitem[Gupta et~al.(2024)Gupta, Yu, Sohn, Gu, Hahn, Li, Essa, Jiang, and Lezama]{gupta2023photorealistic}
Agrim Gupta, Lijun Yu, Kihyuk Sohn, Xiuye Gu, Meera Hahn, Fei-Fei Li, Irfan Essa, Lu Jiang, and Jos{\'e} Lezama.
\newblock Photorealistic video generation with diffusion models.
\newblock \emph{arXiv preprint arXiv:2312.06662}, 2024.

\bibitem[Han et~al.(2022)Han, Xie, and Zisserman]{han2022temporal}
Tengda Han, Weidi Xie, and Andrew Zisserman.
\newblock Temporal alignment networks for long-term video.
\newblock In \emph{CVPR}, 2022.

\bibitem[He et~al.(2022)He, Yang, Zhang, Shan, and Chen]{he2022latent}
Yingqing He, Tianyu Yang, Yong Zhang, Ying Shan, and Qifeng Chen.
\newblock Latent video diffusion models for high-fidelity long video generation.
\newblock \emph{arXiv preprint arXiv:2211.13221}, 2022.

\bibitem[Ho et~al.(2020)Ho, Jain, and Abbeel]{ho2020denoising}
Jonathan Ho, Ajay Jain, and Pieter Abbeel.
\newblock Denoising diffusion probabilistic models.
\newblock \emph{NeurIPS}, 2020.

\bibitem[Ho et~al.(2022{\natexlab{a}})Ho, Chan, Saharia, Whang, Gao, Gritsenko, Kingma, Poole, Norouzi, Fleet, et~al.]{ho2022imagen}
Jonathan Ho, William Chan, Chitwan Saharia, Jay Whang, Ruiqi Gao, Alexey Gritsenko, Diederik~P Kingma, Ben Poole, Mohammad Norouzi, David~J Fleet, et~al.
\newblock Imagen video: High definition video generation with diffusion models.
\newblock \emph{arXiv preprint arXiv:2210.02303}, 2022{\natexlab{a}}.

\bibitem[Ho et~al.(2022{\natexlab{b}})Ho, Salimans, Gritsenko, Chan, Norouzi, and Fleet]{ho2022video}
Jonathan Ho, Tim Salimans, Alexey Gritsenko, William Chan, Mohammad Norouzi, and David~J Fleet.
\newblock Video diffusion models.
\newblock \emph{NeurIPS}, 2022{\natexlab{b}}.

\bibitem[Hu et~al.(2022)Hu, Luo, and Chen]{hu2022make}
Yaosi Hu, Chong Luo, and Zhenzhong Chen.
\newblock Make it move: controllable image-to-video generation with text descriptions.
\newblock In \emph{CVPR}, 2022.

\bibitem[Huberman-Spiegelglas et~al.(2024)Huberman-Spiegelglas, Kulikov, and Michaeli]{huberman2024edit}
Inbar Huberman-Spiegelglas, Vladimir Kulikov, and Tomer Michaeli.
\newblock An edit friendly ddpm noise space: Inversion and manipulations.
\newblock In \emph{CVPR}, 2024.

\bibitem[Kang and Kuo(2024)]{kang2024incorporating}
Xuhui Kang and Yen-Ling Kuo.
\newblock Incorporating task progress knowledge for subgoal generation in robotic manipulation through image edits.
\newblock \emph{arXiv preprint arXiv:2410.11013}, 2024.

\bibitem[Kojima et~al.(2022)Kojima, Gu, Reid, Matsuo, and Iwasawa]{NEURIPS2022_8bb0d291}
Takeshi Kojima, Shixiang~(Shane) Gu, Machel Reid, Yutaka Matsuo, and Yusuke Iwasawa.
\newblock Large language models are zero-shot reasoners.
\newblock In \emph{NeurIPS}, 2022.

\bibitem[Kondratyuk et~al.(2023)Kondratyuk, Yu, Gu, Lezama, Huang, Schindler, Hornung, Birodkar, Yan, Chiu, et~al.]{kondratyuk2023videopoet}
Dan Kondratyuk, Lijun Yu, Xiuye Gu, Jos{\'e} Lezama, Jonathan Huang, Grant Schindler, Rachel Hornung, Vighnesh Birodkar, Jimmy Yan, Ming-Chang Chiu, et~al.
\newblock Videopoet: A large language model for zero-shot video generation.
\newblock \emph{arXiv preprint arXiv:2312.14125}, 2023.

\bibitem[Krojer et~al.(2024)Krojer, Vattikonda, Lara, Jampani, Portelance, Pal, and Reddy]{krojer2024aurora}
Benno Krojer, Dheeraj Vattikonda, Luis Lara, Varun Jampani, Eva Portelance, Christopher Pal, and Siva Reddy.
\newblock Learning action and reasoning-centric image editing from videos and simulations.
\newblock \emph{arXiv preprint arXiv:2407.03471}, 2024.

\bibitem[Lai et~al.(2023)Lai, Dai, Chen, Pang, Rehg, and Liu]{lai2023lego}
Bolin Lai, Xiaoliang Dai, Lawrence Chen, Guan Pang, James~M Rehg, and Miao Liu.
\newblock Lego: Learning egocentric action frame generation via visual instruction tuning.
\newblock \emph{arXiv preprint arXiv:2312.03849}, 2023.

\bibitem[Li et~al.(2024)Li, Chen, Han, Zhang, Wang, and Xie]{li2024multi}
Zeqian Li, Qirui Chen, Tengda Han, Ya Zhang, Yanfeng Wang, and Weidi Xie.
\newblock Multi-sentence grounding for long-term instructional video.
\newblock 2024.

\bibitem[Liang et~al.(2024)Liang, Liu, Ozguroglu, Sudhakar, Dave, Tokmakov, Song, and Vondrick]{liang2024dreamitate}
Junbang Liang, Ruoshi Liu, Ege Ozguroglu, Sruthi Sudhakar, Achal Dave, Pavel Tokmakov, Shuran Song, and Carl Vondrick.
\newblock Dreamitate: Real-world visuomotor policy learning via video generation.
\newblock \emph{arXiv preprint arXiv:2406.16862}, 2024.

\bibitem[Mavroudi et~al.(2023)Mavroudi, Afouras, and Torresani]{mavroudi2023learning}
Effrosyni Mavroudi, Triantafyllos Afouras, and Lorenzo Torresani.
\newblock Learning to ground instructional articles in videos through narrations.
\newblock In \emph{ICCV}, 2023.

\bibitem[Mei and Patel(2023)]{mei2023vidm}
Kangfu Mei and Vishal Patel.
\newblock Vidm: Video implicit diffusion models.
\newblock In \emph{AAAI}, 2023.

\bibitem[Menon et~al.(2024)Menon, Misra, and Girdhar]{menon2024generating}
Sachit Menon, Ishan Misra, and Rohit Girdhar.
\newblock Generating illustrated instructions.
\newblock In \emph{CVPR}, 2024.

\bibitem[Miech et~al.(2019)Miech, Zhukov, Alayrac, Tapaswi, Laptev, and Sivic]{miech2019howto100m}
Antoine Miech, Dimitri Zhukov, Jean-Baptiste Alayrac, Makarand Tapaswi, Ivan Laptev, and Josef Sivic.
\newblock Howto100m: Learning a text-video embedding by watching hundred million narrated video clips.
\newblock In \emph{ICCV}, 2019.

\bibitem[Nair et~al.(2018)Nair, Pong, Dalal, Bahl, Lin, and Levine]{nair2018visual}
Ashvin~V Nair, Vitchyr Pong, Murtaza Dalal, Shikhar Bahl, Steven Lin, and Sergey Levine.
\newblock Visual reinforcement learning with imagined goals.
\newblock \emph{NeurIPS}, 2018.

\bibitem[Oquab et~al.(2024)Oquab, Darcet, Moutakanni, Vo, Szafraniec, Khalidov, Fernandez, HAZIZA, Massa, El-Nouby, et~al.]{oquab2023dinov2}
Maxime Oquab, Timoth{\'e}e Darcet, Th{\'e}o Moutakanni, Huy~V Vo, Marc Szafraniec, Vasil Khalidov, Pierre Fernandez, Daniel HAZIZA, Francisco Massa, Alaaeldin El-Nouby, et~al.
\newblock Dinov2: Learning robust visual features without supervision.
\newblock \emph{TMLR}, 2024.

\bibitem[Phung et~al.(2024)Phung, Ge, and Huang]{phung2024coherent}
Quynh Phung, Songwei Ge, and Jia-Bin Huang.
\newblock Coherent zero-shot visual instruction generation.
\newblock \emph{arXiv preprint arXiv:2406.04337}, 2024.

\bibitem[Polyak et~al.(2024)Polyak, Zohar, Brown, Tjandra, Sinha, Lee, Vyas, Shi, Ma, Chuang, et~al.]{polyak2024movie}
Adam Polyak, Amit Zohar, Andrew Brown, Andros Tjandra, Animesh Sinha, Ann Lee, Apoorv Vyas, Bowen Shi, Chih-Yao Ma, Ching-Yao Chuang, et~al.
\newblock Movie gen: A cast of media foundation models.
\newblock \emph{arXiv preprint arXiv:2410.13720}, 2024.

\bibitem[Radford et~al.(2021)Radford, Kim, Hallacy, Ramesh, Goh, Agarwal, Sastry, Askell, Mishkin, Clark, et~al.]{radford2021learning}
Alec Radford, Jong~Wook Kim, Chris Hallacy, Aditya Ramesh, Gabriel Goh, Sandhini Agarwal, Girish Sastry, Amanda Askell, Pamela Mishkin, Jack Clark, et~al.
\newblock Learning transferable visual models from natural language supervision.
\newblock In \emph{ICML}, 2021.

\bibitem[Rombach et~al.(2022)Rombach, Blattmann, Lorenz, Esser, and Ommer]{rombach2022high}
Robin Rombach, Andreas Blattmann, Dominik Lorenz, Patrick Esser, and Bj{\"o}rn Ommer.
\newblock High-resolution image synthesis with latent diffusion models.
\newblock In \emph{CVPR}, 2022.

\bibitem[RunwayML(2024)]{runwayml2024gen3}
RunwayML.
\newblock Gen-3 alpha.
\newblock 2024.

\bibitem[Shvetsova et~al.(2024)Shvetsova, Kukleva, Hong, Rupprecht, Schiele, and Kuehne]{shvetsova2025howtocaption}
Nina Shvetsova, Anna Kukleva, Xudong Hong, Christian Rupprecht, Bernt Schiele, and Hilde Kuehne.
\newblock Howtocaption: Prompting llms to transform video annotations at scale.
\newblock In \emph{ECCV}, 2024.

\bibitem[Song et~al.(2024)Song, Byrne, Nagarajan, Wang, Martin, and Torresani]{song2024ego4d}
Yale Song, Eugene Byrne, Tushar Nagarajan, Huiyu Wang, Miguel Martin, and Lorenzo Torresani.
\newblock Ego4d goal-step: Toward hierarchical understanding of procedural activities.
\newblock \emph{NeurIPS}, 2024.

\bibitem[Sou\v{c}ek et~al.(2024{\natexlab{a}})Sou\v{c}ek, Alayrac, Miech, Laptev, and Sivic]{soucek2024multitask}
Tom\'{a}\v{s} Sou\v{c}ek, Jean-Baptiste Alayrac, Antoine Miech, Ivan Laptev, and Josef Sivic.
\newblock Multi-task learning of object states and state-modifying actions from web videos.
\newblock \emph{TPAMI}, 2024{\natexlab{a}}.

\bibitem[Sou\v{c}ek et~al.(2024{\natexlab{b}})Sou\v{c}ek, Damen, Wray, Laptev, and Sivic]{soucek2024genhowto}
Tom\'{a}\v{s} Sou\v{c}ek, Dima Damen, Michael Wray, Ivan Laptev, and Josef Sivic.
\newblock Genhowto: Learning to generate actions and state transformations from instructional videos.
\newblock In \emph{CVPR}, 2024{\natexlab{b}}.

\bibitem[Tang et~al.(2019)Tang, Ding, Rao, Zheng, Zhang, Zhao, Lu, and Zhou]{tang2019coin}
Yansong Tang, Dajun Ding, Yongming Rao, Yu Zheng, Danyang Zhang, Lili Zhao, Jiwen Lu, and Jie Zhou.
\newblock Coin: A large-scale dataset for comprehensive instructional video analysis.
\newblock In \emph{CVPR}, 2019.

\bibitem[Wang et~al.(2024)Wang, Yuan, Zhang, Chen, Wang, Zhang, Shen, Zhao, and Zhou]{wang2024videocomposer}
Xiang Wang, Hangjie Yuan, Shiwei Zhang, Dayou Chen, Jiuniu Wang, Yingya Zhang, Yujun Shen, Deli Zhao, and Jingren Zhou.
\newblock Videocomposer: Compositional video synthesis with motion controllability.
\newblock \emph{NeurIPS}, 2024.

\bibitem[Xing et~al.(2024)Xing, Xia, Zhang, Chen, Yu, Liu, Liu, Wang, Shan, and Wong]{xing2024dynamicrafter}
Jinbo Xing, Menghan Xia, Yong Zhang, Haoxin Chen, Wangbo Yu, Hanyuan Liu, Gongye Liu, Xintao Wang, Ying Shan, and Tien-Tsin Wong.
\newblock Dynamicrafter: Animating open-domain images with video diffusion priors.
\newblock In \emph{ECCV}, 2024.

\bibitem[Xu et~al.(2024)Xu, Xu, Xu, Chi, Wetzstein, Veloso, and Song]{xu2024flow}
Mengda Xu, Zhenjia Xu, Yinghao Xu, Cheng Chi, Gordon Wetzstein, Manuela Veloso, and Shuran Song.
\newblock Flow as the cross-domain manipulation interface.
\newblock \emph{arXiv preprint arXiv:2407.15208}, 2024.

\bibitem[Xue et~al.(2024)Xue, Ashutosh, and Grauman]{xue2024learning}
Zihui Xue, Kumar Ashutosh, and Kristen Grauman.
\newblock Learning object state changes in videos: An open-world perspective.
\newblock In \emph{CVPR}, 2024.

\bibitem[Yan et~al.(2023)Yan, Xiong, Nagrani, Arnab, Wang, Ge, Ross, and Schmid]{yan2023unloc}
Shen Yan, Xuehan Xiong, Arsha Nagrani, Anurag Arnab, Zhonghao Wang, Weina Ge, David Ross, and Cordelia Schmid.
\newblock Unloc: A unified framework for video localization tasks.
\newblock In \emph{ICCV}, 2023.

\bibitem[Yang et~al.(2023)Yang, Srivastava, and Mandt]{yang2023diffusion}
Ruihan Yang, Prakhar Srivastava, and Stephan Mandt.
\newblock Diffusion probabilistic modeling for video generation.
\newblock \emph{Entropy}, 2023.

\bibitem[Yang et~al.(2024{\natexlab{a}})Yang, Du, Ghasemipour, Tompson, Kaelbling, Schuurmans, and Abbeel]{yang2024learning}
Sherry Yang, Yilun Du, Seyed Kamyar~Seyed Ghasemipour, Jonathan Tompson, Leslie~Pack Kaelbling, Dale Schuurmans, and Pieter Abbeel.
\newblock Learning interactive real-world simulators.
\newblock In \emph{ICLR}, 2024{\natexlab{a}}.

\bibitem[Yang et~al.(2021)Yang, Panagopoulou, Lyu, Zhang, Yatskar, and Callison-Burch]{yang2021visual}
Yue Yang, Artemis Panagopoulou, Qing Lyu, Li Zhang, Mark Yatskar, and Chris Callison-Burch.
\newblock Visual goal-step inference using wikihow.
\newblock In \emph{EMNLP}, 2021.

\bibitem[Yang et~al.(2024{\natexlab{b}})Yang, Teng, Zheng, Ding, Huang, Xu, Yang, Hong, Zhang, Feng, et~al.]{yang2024cogvideox}
Zhuoyi Yang, Jiayan Teng, Wendi Zheng, Ming Ding, Shiyu Huang, Jiazheng Xu, Yuanming Yang, Wenyi Hong, Xiaohan Zhang, Guanyu Feng, et~al.
\newblock Cogvideox: Text-to-video diffusion models with an expert transformer.
\newblock \emph{arXiv preprint arXiv:2408.06072}, 2024{\natexlab{b}}.

\bibitem[Yu et~al.(2023{\natexlab{a}})Yu, Sohn, Kim, and Shin]{yu2023video}
Sihyun Yu, Kihyuk Sohn, Subin Kim, and Jinwoo Shin.
\newblock Video probabilistic diffusion models in projected latent space.
\newblock In \emph{CVPR}, 2023{\natexlab{a}}.

\bibitem[Yu et~al.(2023{\natexlab{b}})Yu, Xiao, Stone, Tompson, Brohan, Wang, Singh, Tan, Peralta, Ichter, et~al.]{yu2023scaling}
Tianhe Yu, Ted Xiao, Austin Stone, Jonathan Tompson, Anthony Brohan, Su Wang, Jaspiar Singh, Clayton Tan, Jodilyn Peralta, Brian Ichter, et~al.
\newblock Scaling robot learning with semantically imagined experience.
\newblock \emph{arXiv preprint arXiv:2302.11550}, 2023{\natexlab{b}}.

\bibitem[Zhai et~al.(2023)Zhai, Mustafa, Kolesnikov, and Beyer]{zhai2023sigmoid}
Xiaohua Zhai, Basil Mustafa, Alexander Kolesnikov, and Lucas Beyer.
\newblock Sigmoid loss for language image pre-training.
\newblock In \emph{ICCV}, 2023.

\bibitem[Zhang et~al.(2024)Zhang, Wu, Liu, Zhao, Ran, Gu, Gao, and Shou]{zhang2024show}
David~Junhao Zhang, Jay~Zhangjie Wu, Jia-Wei Liu, Rui Zhao, Lingmin Ran, Yuchao Gu, Difei Gao, and Mike~Zheng Shou.
\newblock Show-1: Marrying pixel and latent diffusion models for text-to-video generation.
\newblock \emph{IJCV}, 2024.

\bibitem[Zhang et~al.(2023)Zhang, Wang, Zhang, Zhao, Yuan, Qin, Wang, Zhao, and Zhou]{zhang2023i2vgen}
Shiwei Zhang, Jiayu Wang, Yingya Zhang, Kang Zhao, Hangjie Yuan, Zhiwu Qin, Xiang Wang, Deli Zhao, and Jingren Zhou.
\newblock I2vgen-xl: High-quality image-to-video synthesis via cascaded diffusion models.
\newblock \emph{arXiv preprint arXiv:2311.04145}, 2023.

\bibitem[Zhou et~al.(2022)Zhou, Wang, Yan, Lv, Zhu, and Feng]{zhou2022magicvideo}
Daquan Zhou, Weimin Wang, Hanshu Yan, Weiwei Lv, Yizhe Zhu, and Jiashi Feng.
\newblock Magicvideo: Efficient video generation with latent diffusion models.
\newblock \emph{arXiv preprint arXiv:2211.11018}, 2022.

\bibitem[Zhukov et~al.(2019)Zhukov, Alayrac, Cinbis, Fouhey, Laptev, and Sivic]{zhukov2019cross}
Dimitri Zhukov, Jean-Baptiste Alayrac, Ramazan~Gokberk Cinbis, David Fouhey, Ivan Laptev, and Josef Sivic.
\newblock Cross-task weakly supervised learning from instructional videos.
\newblock In \emph{CVPR}, 2019.

\end{thebibliography}
\flushcolsend
}

\ARXIVversion{\newpage\appendix\ARXIVversion{\section*{Appendix}}{\section*{Overview}}

In the \ARXIVversion{appendix}{supplementary material}, we first provide dataset collection details and show examples from our dataset in Section~\ref{supmatsec:dscolldetails}. Then, in Section~\ref{supmatsec:dscomparison}, we compare our dataset to other related works and discuss their differences. We provide implementation details in Section~\ref{supmatsec:impldetails} and evaluation details and analysis of our metrics in Section~\ref{supmatsec:evaldetails}. In Section~\ref{supmatsec:addquantres}, additional quantitative results are provided, and in Section~\ref{supmatsec:qualres}, we show a large variety of qualitative results.

\section{Dataset Collection Details}\label{supmatsec:dscolldetails}

\paragraphcustomWOvspace{Speech transcription: YouTube ASR vs. WhisperX.}
The original video transcripts released with the HowTo100M dataset, generated through YouTube ASR, are known to contain errors~\cite{han2022temporal,li2024multi}. Following Li \etal~\cite{li2024multi}, we employ WhisperX~\cite{bain2021frozen} to obtain higher-quality transcripts for narrated instructional videos. As shown in Figure~\ref{supmatfig:asr_comparison}, WhisperX provides improved transcription quality over YouTube ASR through better punctuation, fewer transcription errors (\eg, \textit{``Put some aluminum foil in there''} vs. YouTube ASR's incorrect \textit{``it's a moment of oil in there''}), and more accurate sentence segmentation with timestamps. These improvements are essential for our subsequent dataset creation steps, which rely on accurate transcription.

\paragraphcustom{Details on filtering of irrelevant videos.} To address the presence of non-instructional content in HowTo100M, we prompt Llama 3~\cite{dubey2024llama} (\texttt{Llama\--3.1-8B-Instruct}) to identify whether videos are instructional or not based on their transcript excerpts. The full prompt used is shown in Figure~\ref{supmatfig:prompt-filtering}.

We validate our approach for video filtering by manually annotating a balanced validation set of 200 videos. Our approach achieves the false positive rate of 5\% and the false negative rate of 12\%, \ie, 95\% of non-instructional videos spuriously present in the HowTo100M dataset are filtered out by our automatic approach.
Applied to the full HowTo100M dataset, we identified 847K (68.4\%) instructional videos and filtered out 391K (31.6\%) non-instructional content, including product reviews (\eg, \textit{``Houseplant Unboxing | Steve’s Leaves''}), entertainment videos (\eg, \textit{``Don McLean - American Pie (with Lyrics)''}), and personal vlogs (\eg, \textit{``Driving to the West, an RV lifestyle vlog''}).

\begin{figure}
    \centering
    \includegraphics[width=\linewidth]{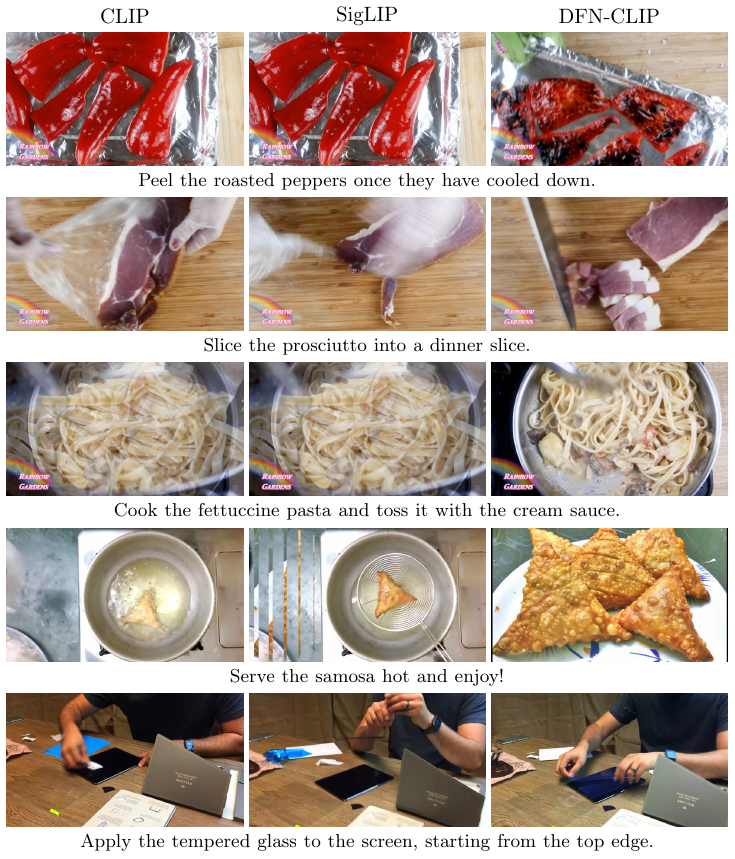}
    \caption{\textbf{Frame matching comparison across CLIP, SigLIP, and DFN-CLIP (used in our work).} For each method, the figure shows the best matching frame to the instructional text (shown below).}
    \label{supmatfig:clip_comparison}
\end{figure}

\begin{figure*}[t]
    \centering
    \includegraphics[width=\linewidth]{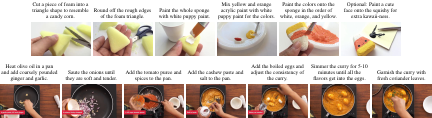}
    \caption{\textbf{Samples from the ShowHowTo dataset}. Each sample (row) is a sequence of textual instructions (top) and the associated visual instruction images (bottom).}
    \label{supmatfig:dsexample}
\end{figure*}

\begin{figure*}[ht]
    \definecolor{greenok}{RGB}{0,158,96}

    \begin{minipage}{0.49\textwidth}
    {\scriptsize
\begin{tabular}{r@{\hskip 0.15cm}c@{\hskip 0.15cm}l}
18.05 - 19.53 & \textcolor{greenok}{\cmark} & Put aluminum foil in the pressure cooker.\\
28.86 - 29.70 & \textcolor{greenok}{\cmark} & Create two little rings using the aluminum foil.\\
38.50 - 40.42 & \textcolor{greenok}{\cmark} & Place the potatoes in the pressure cooker, above the water.\\
44.99 - 50.23 & \textcolor{greenok}{\cmark} & Add 1.5 cups of water to the pressure cooker.\\
59.85 - 62.05 & \textcolor{greenok}{\cmark} & Cook the potatoes for 15 minutes, depending on their size.\\
76.08 - 80.48 & \textcolor{cvprblue}{\textbf{?}} & Remove the potatoes from the pressure cooker and serve.\\
\end{tabular}
}
\end{minipage}
\hfill
\begin{minipage}{0.49\textwidth}
{\scriptsize
\begin{tabular}{r@{\hskip 0.15cm}c@{\hskip 0.15cm}l}
 4.00 -  12.00 & \textcolor{cvprblue}{\textbf{?}} & Put potatoes in the pressure cooker and create steaming rings.\\
17.00 -  25.00 & \textcolor{cvprblue}{\textbf{?}} & Add one and a half cups of water to the pressure cooker.\\
19.00 -  27.00 & \textcolor{greenok}{\cmark} & Place aluminum foil in the pressure cooker.\\
78.00 -  86.00 & \textcolor{red}{\xmark} & It doesn't heat up your kitchen.\\
83.00 -  91.00 & \textcolor{red}{\xmark} & It's really good for in the summer when you want to have\\
               & & something like a baked potato.\\
86.00 -  94.00 & \textcolor{red}{\xmark} & Up anything else.\\
91.00 -  99.00 & \textcolor{red}{\xmark} & When they're done, you just pop\\
96.00 - 104.00 & \textcolor{red}{\xmark} & It's just going to be steamed.\\
96.00 - 104.00 & \textcolor{red}{\xmark} & So there it is.\\
\end{tabular}
}
\end{minipage}
\vspace{0.2cm}
\begin{minipage}{\textwidth}
\centering
{\footnotesize
\begin{tabular}{ccc}
\textcolor{greenok}{\cmark} -- \textit{an instruction with correct approximate timestamp} & \textcolor{cvprblue}{\textbf{?}} -- \textit{an instruction with incorrect timestamp} & \textcolor{red}{\xmark} -- \textit{not an instruction} \\
\end{tabular}
}
\end{minipage}
% this is the video https://www.youtube.com/watch?v=rfJZFahAdPc

    \vspace*{-7mm}
    \caption{\textbf{Comparison between textual instructions} extracted by our method (left) and the textual instructions from HowToStep~\cite{li2024multi} (right) for the same randomly chosen \textit{`How to bake a potato in the pressure cooker'} video. The original transcript used to produce our instructions is shown in Figure~\ref{supmatfig:asr_comparison}. Our method correctly identifies the key steps in the narrations and summarizes them in step-by-step instructions. On the other hand, the HowToStep data often contain steps that are not instructions.}
    \label{supmatfig:extracted_step_comparison}
\end{figure*}

\paragraphcustom{Step extraction details.}
\added{Our approach uses Llama 3 to extract the steps from video narrations. This contrasts the related work \cite{soucek2024genhowto}, which extracts steps from video frames and uses an image-captioning model to generate step captions. We found that the latter approach resulted in captions that were too brief, high-level, and lacked sufficient detail to differentiate adjacent steps.}

Figure~\ref{supmatfig:prompt-step-extraction} illustrates our Llama 3 prompt for extracting instructional steps from video narrations, including the few-shot examples used (truncated to fit into a page; complete prompts \ARXIVversion{are available on the project website}{will be released with our code}). We processed videos under 10 minutes long, as longer transcripts exceeded the context limit of Llama 3 and also degraded output quality. The model is instructed to generate temporally ordered steps with approximate timestamps using video transcripts. We filter out malformed results with non-temporally ordered steps or incomplete descriptions.
Examples of extracted steps are shown in Figure~\ref{supmatfig:dsexample}.

\paragraphcustom{Cross-modal frame alignment details.} For each extracted step, we find its matching frame in the video using DFN-CLIP~\cite{fang2023data} (\texttt{DFN5B-CLIP-ViT-H-14-378}), restricting the search to frames within the step's time bounds generated by Llama 3. We formulate this as a dynamic programming problem to find optimal frame-text pairs while preserving temporal order and maximizing alignment scores.

\added{Although the temporal boundaries generated by Llama~3 are rarely incorrect, narrations do not always align with respect to video frames and can occur slightly before/after the visuals.} To account for the well-known temporal misalignment~\cite{han2022temporal, li2024multi}, we expand the temporal boundaries by a fixed duration of $\epsilon$ seconds, increasing the search space for frame alignment. Through analysis on a small validation set of manually annotated videos with precise step boundaries, we found $\epsilon=15$ seconds yielded a good balance of precision (the correct frame was selected by DFN-CLIP) and recall (the correct frame was inside the searched interval).

We use DFN-CLIP over related contrastive models such as CLIP~\cite{radford2021learning} and SigLIP~\cite{zhai2023sigmoid} as we found it superior in certain cases for matching video frames to instructional steps.
We observed CLIP and SigLIP often exhibited limitations such as incorrect object state identification (\eg, unroasted peppers, uncut prosciutto in Figure~\ref{supmatfig:clip_comparison}, rows 1-2) and tendency to select blurry or transition frames (Figure~\ref{supmatfig:clip_comparison}, rows 2-3).
\added{We quantify the quality of the automatically selected keyframes by a small user study. We manually annotated keyframes for 100 steps and asked humans to blindly select whether the manually or automatically selected frame better corresponds to the text instruction. 
DFN-CLIP was preferred in 18\% of cases, human annotation was preferred in 36\% of cases, remaining 46\% of cases were a tie. This indicates fairly decent alignment with human judgment.}

\paragraphcustom{Dataset Statistics.}
We analyze the ShowHowTo dataset statistics in Figures~\ref{supmatfig:dataset-statistics} and~\ref{supmatfig:word-clouds}. The dataset encompasses 25K tasks across diverse categories, including Food and Entertainment, Hobbies and Crafts, \etc., derived from HowTo100M's task hierarchy. Figure~\ref{supmatfig:dataset-statistics} (left) shows the distribution of these categories in our dataset. Each sample contains an average of 7.7 steps, with 11.37 words per step (Figure~\ref{supmatfig:dataset-statistics}, right). The word clouds in Figure~\ref{supmatfig:word-clouds} showcases common verbs of physical actions like \textit{remove}, \textit{add}, and \textit{make}, alongside various household objects and materials used in everyday tasks.

\begin{figure*}[t]
    \centering
    \begin{subfigure}[b]{0.65\textwidth}
        \centering
        \includegraphics[width=\textwidth]{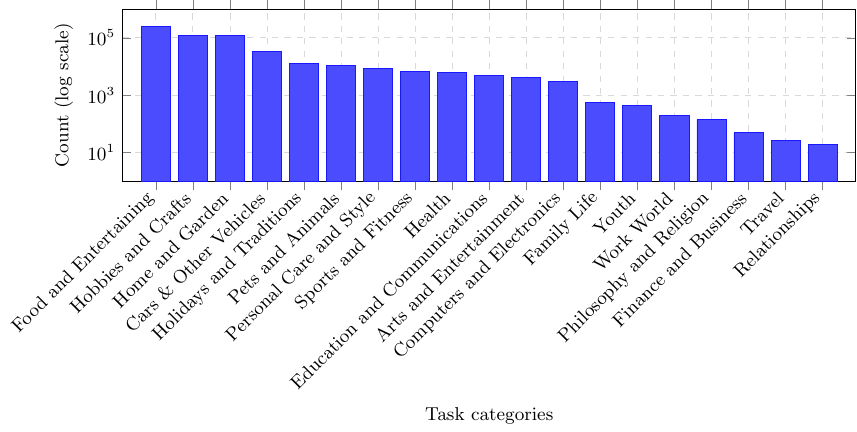}
        \label{supmatfig:dataset-statistics-categories}
    \end{subfigure}
    \hfill
    \begin{minipage}[b]{0.3\textwidth}
        \begin{subfigure}[b]{\textwidth}
            \centering
            \includegraphics[width=\textwidth]{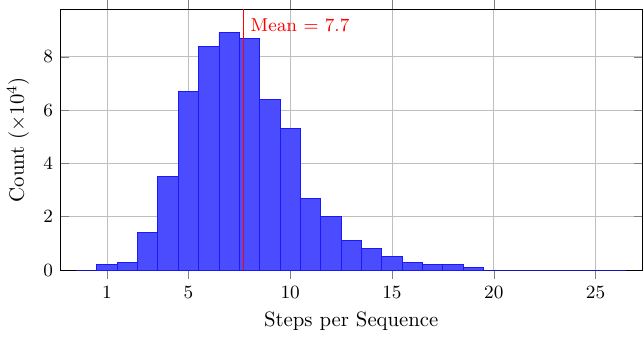}
            \label{supmatfig:dataset-statistics-steps}
        \end{subfigure}
        % \vspace{0.3cm}
        \begin{subfigure}[b]{\textwidth}
            \centering
            \includegraphics[width=\textwidth]{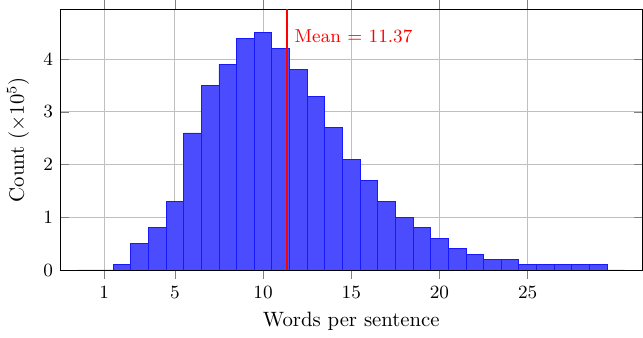}
            \label{supmatfig:dataset-statistics-words}
        \end{subfigure}
    \end{minipage}
    \vspace{-0.4cm}
    \caption{\textbf{Statistics of the ShowHowTo dataset}. Left: Distribution of task categories in ShowHowTo dataset. Top-right: Distribution of the number of steps per sequence. Bottom-right: Distribution of the number of words per sentence.}
    \label{supmatfig:dataset-statistics}
\end{figure*}

\begin{figure*}[t]
    \centering
    \begin{subfigure}[b]{0.48\textwidth}
        \centering
        \includegraphics[width=\textwidth]{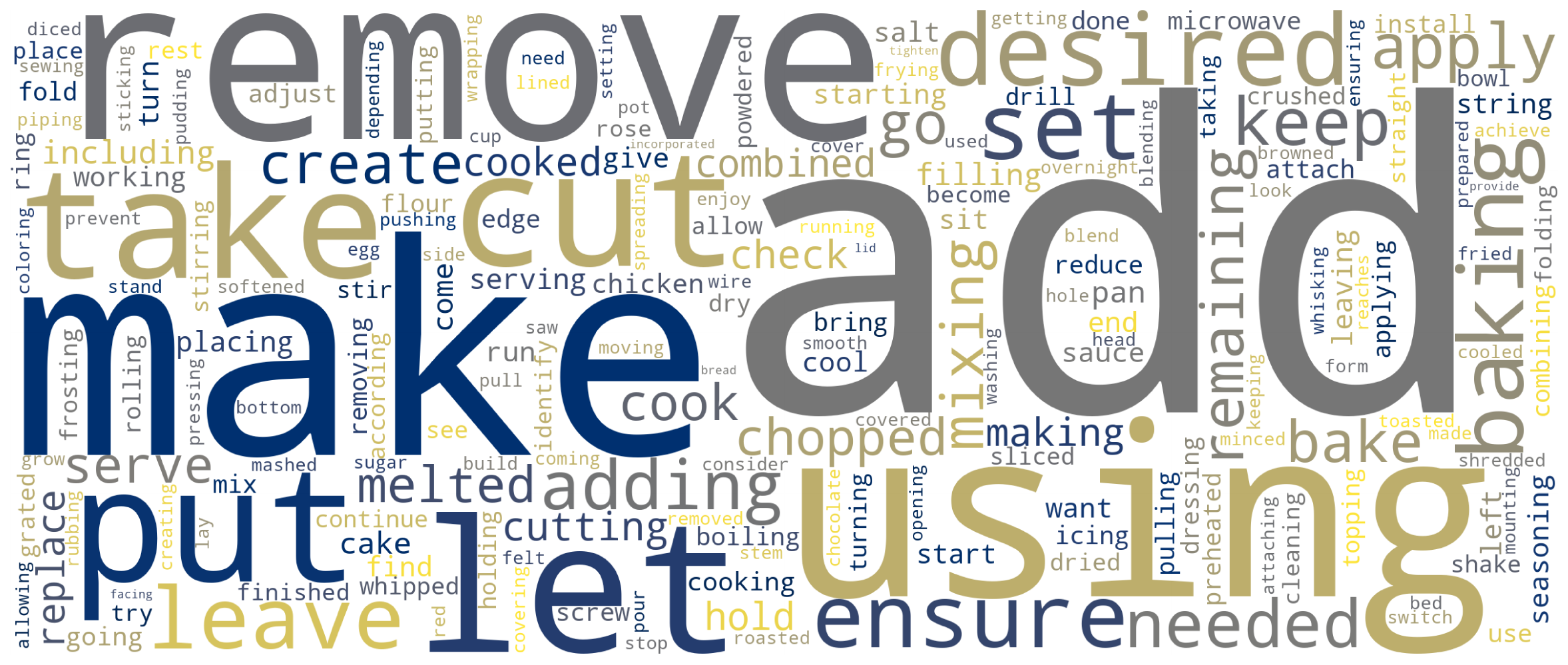}
    \end{subfigure}
    \hfill
    \begin{subfigure}[b]{0.48\textwidth}
        \centering
        \includegraphics[width=\textwidth]{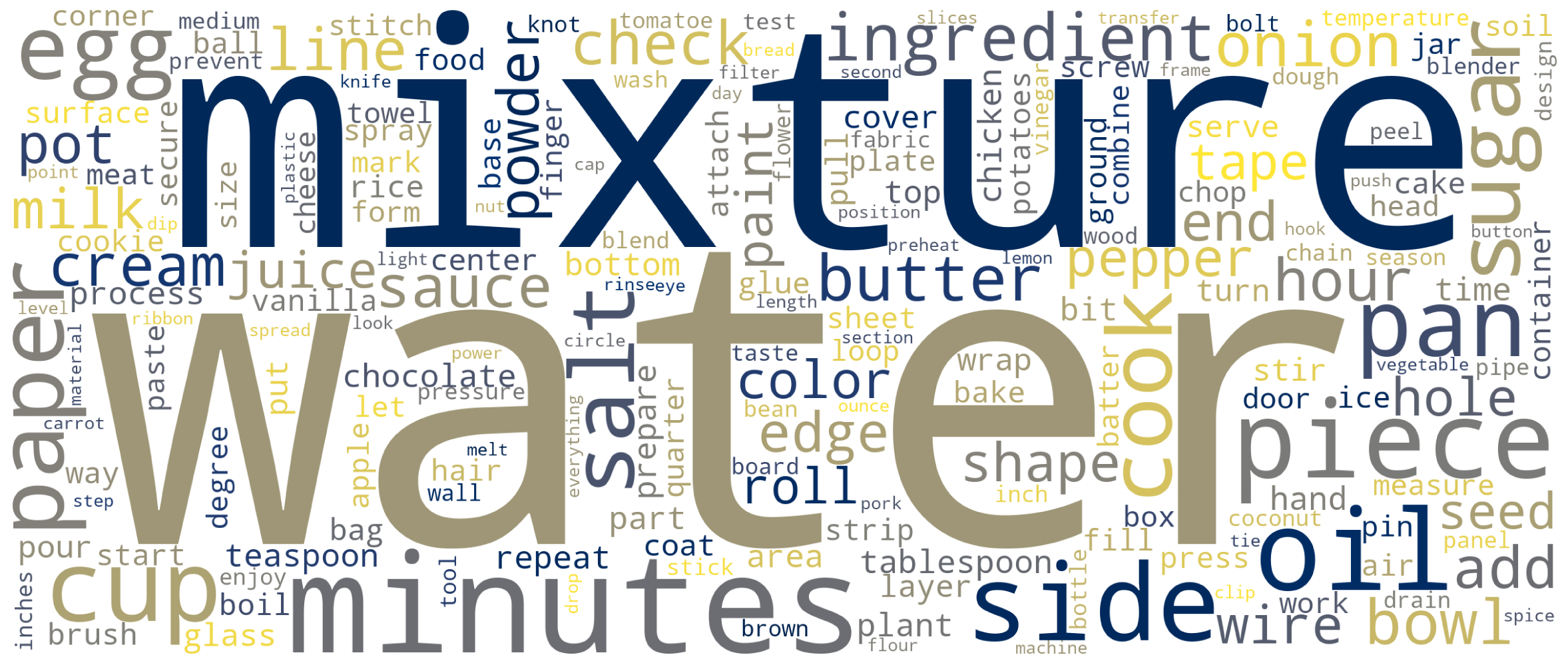}
    \end{subfigure}
    \caption{\textbf{Word cloud visualizations} of verbs (left) and nouns (right) in the textual instructions of ShowHowTo dataset.}
    \label{supmatfig:word-clouds}
\end{figure*}

\begin{table*}[t]
\centering
\scriptsize
\begin{tabular}{lccccccccc}
\toprule
\multirow{2}{*}{Dataset} & \multirow{2}{*}{Source} & Manually & Task & Visual & \multirow{2}{*}{Scale} & \multirow{2}{*}{\# Tasks} & Avg.& \multirow{2}{*}{Visual Type}  & \multirow{2}{*}{Annotation Type} \\
& & Curated &Domain &Domain & & & Steps / Seq. & & \\
\midrule
CrossTask~\cite{zhukov2019cross} & YouTube & \cmark & Open & Open & 4.7K & 18 & 7.4 & Video Segments & Categorical \\
COIN~\cite{tang2019coin} & YouTube & \cmark & Open & Open & 10K & 180 & 3.9 & Video Segments & Categorical \\
Ego4D Goal-Step~\cite{song2024ego4d} & Ego4D & \cmark & Open & Egocentric & 717\textsuperscript{$\ddagger$} & 80 & 23.3 & Video Segments & Instructions \\
% HTMAlign~\cite{han2022temporal} & HowTo100M & Open & 1.2M & Video Segments & - & Narrations \\
LEGO~\cite{lai2023lego} & Ego4D \& EPIC-K. & \xmark & Open & Egocentric & 147K & - & 1.0 & Key Frames & Instructions \\
AURORA~\cite{krojer2024aurora} & Multiple & \cmark & Open & Open & 289K & - & 1.0 & Key Frames & Instructions \\
GenHowTo~\cite{soucek2024genhowto} & COIN \& ChangeIt & \xmark & Open & Open & 45K & 224 & 2.0 & Key Frames & Captions \\
HowToCaption~\cite{shvetsova2025howtocaption} & HowTo100M & \xmark & Open & Open & 1.1M & 23.6K & 18.5 & Video Segments & Captions\\
HT-Step~\cite{afouras2024ht} & HowTo100M & \cmark & Cooking & Open & 18K & 433 & 5.9 & Video Segments & Instructions\\
HowToStep~\cite{li2024multi} & HowTo100M & \xmark & Cooking & Open & 312K & 14.2K & 10.6 & Video Segments & Instructions \\
WikiHow-VGSI~\cite{yang2021visual} & WikiHow & \cmark & Open & Illustrations\textsuperscript{$\dagger$} & 100K & 53.2K & 6.0 & Key Frames & Instructions \\
\midrule
\textbf{ShowHowTo} & HowTo100M & \xmark & Open & Open & 578K & 25K & 7.7 & Key Frames & Instructions \\
\bottomrule
\multicolumn{10}{c}{\textsuperscript{$\dagger$}~{\scriptsize Some examples in the dataset are real photos.}\quad\quad\textsuperscript{$\ddagger$}~{\scriptsize The subset with step annotations.}}\\
\end{tabular}
\caption{\textbf{Comparison of instructional datasets}. Scale refers to the number of instruction sequences of image-text pairs. Annotation Type describes the nature of text captions. Visual Type indicates the format of visual content.}
\label{tab:dataset_comparison}
\end{table*}

\section{Relation to Existing Datasets}\label{supmatsec:dscomparison}
We show the comparison of related instructional datasets in Table~\ref{tab:dataset_comparison}.  Early datasets like CrossTask~\cite{zhukov2019cross} and COIN~\cite{tang2019coin} are manually curated but small in scale with categorical instruction annotations. While recent datasets like HowToCaption~\cite{shvetsova2025howtocaption} have expanded significantly (1.1M sequences), they provide generic captions rather than instructional text annotations. Specialized datasets exist for egocentric domains (LEGO~\cite{lai2023lego}, Ego4D Goal-Step~\cite{song2024ego4d}) and single-step instructions (AURORA~\cite{krojer2024aurora}, GenHowTo~\cite{soucek2024genhowto}), but their narrow scope limits general applicability.

The two most comparable large-scale multi-step instructional datasets to our dataset are WikiHow-VGSI~\cite{yang2021visual} (100K sequences) and HowToStep~\cite{li2024multi} (312K sequences).
WikiHow-VGSI, composed of image-step pairs extracted from WikiHow articles, predominantly contains digitally drawn illustrations rather than real photos, making it unsuitable for realistic image generation and very difficult to scale. On the other hand, HowToStep~\cite{li2024multi}, similarly to us, leverages HowTo100M videos, but its focus is solely on the cooking domain, and it contains only temporal video segments instead of individual representative images.
Lastly, the provided segments are of variable quality, often not being instructional (see Figure~\ref{supmatfig:extracted_step_comparison} for an example). In comparison, the ShowHowTo dataset is both larger in scale and contains more diverse tasks with higher-quality steps. In \ARXIVversion{Section~\ref{subsec:ablations}}{the main paper}, we train our model on both WikiHow-VGSI and HowToStep, showing that training on our dataset results in models with substantially better generation capabilities.

\section{Implementation Details}\label{supmatsec:impldetails}

\paragraphcustomWOvspace{ShowHowTo implementation details.}
We initialize our model with pretrained weights of DynamiCrafter for image animation~\cite{xing2024dynamicrafter} and train the model for approximately 100K steps on four AMD MI250x multi-chip modules (8 GPUs) with a total of 512 GB of VRAM (64 GB per GPU). We use the batch size of 16 image sequences at the resolution of 256$\times$256 pixels. The image sequence length is variable, ranging between two and eight. We train the model using AdamW optimizer with a learning rate of $2\cdot 10^{-5}$. The training takes approximately 48 hours. During inference, we generate visual instruction sequences using the DDIM sampler with 50 denoising steps. Both the inference and training code, along with trained model weights \ARXIVversion{are}{will be made} publicly available\ARXIVversion{ at \url{https://soczech.github.io/showhowto/}}{}.

\paragraphcustom{Implementation details of the related methods.}
In our evaluation of the related methods, we use the official implementations of GenHowTo, AURORA, and InstructPix2Pix provided by the respective authors. In the case of GenHowTo, we use the action model weights. For StackedDiffusion~\cite{menon2024generating}, we use the official implementation modified to replace the first generated image with the noised version of the input image in each denoising step, to enable input image conditioning. The work of Phung \etal~\cite{phung2024coherent} does not provide the official implementation at the time of writing; therefore, we reimplement it using the Stable Diffusion 2.1 model. For the step similarity matrix, we use the matrix with all values set to one, and similarly to the StackedDiffusion model, we replace the first generated image with the noised version of the input image in each denoising step.

\begin{figure}[!ht]
    \centering
    \includegraphics[width=\linewidth]{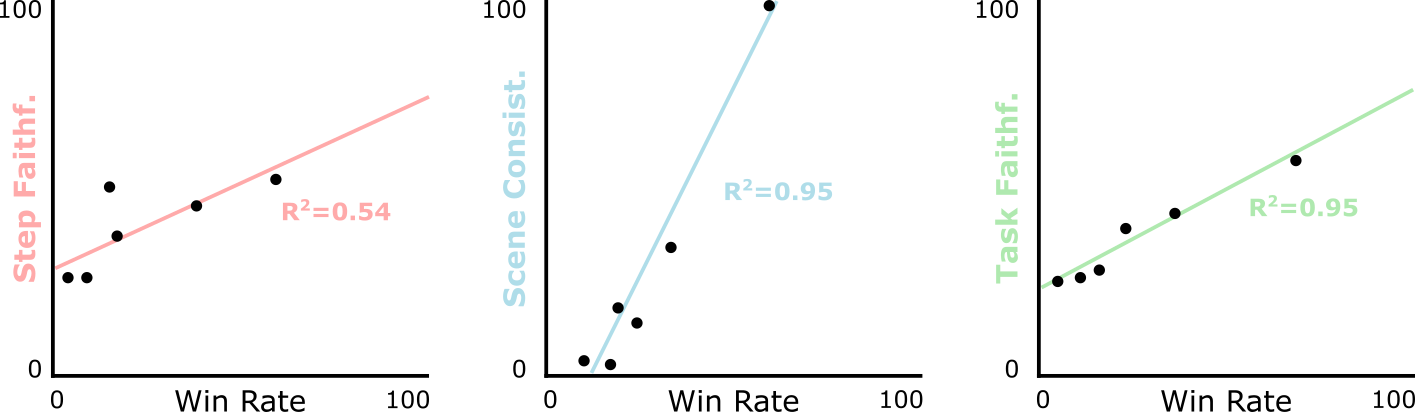}
    \caption{\textbf{Correlation of our metrics with human preference.} User study results (Win Rate \%) highly correlate with our three metrics---Step Faithfulness, Scene Consistency, and Task Faithfulness.}
    \label{fig:user-study-corr}
\end{figure}

\section{Evaluation Details}\label{supmatsec:evaldetails}
As described in \ARXIVversion{Section~\ref{subsec:eval-details}}{the main paper}, we evaluate our method using three metrics: Step Faithfulness, Scene Consistency, and Task Faithfulness. 
DFN-CLIP~\cite{fang2023data} (\texttt{DFN5B\--CLIP\--ViT-H-14-378}) is used for the computation of the Step Faithfulness and Task Faithfulness metrics. Scene Consistency is computed with the averaged spatial patch features of DINOv2~\cite{oquab2023dinov2} (\texttt{dinov2\_vitb14\_reg}). All metrics are first averaged per sequence before being averaged across the test set to account for the variable sequence length.

We verify how well the used metrics correlate with human preference as measured by our user study (see \ARXIVversion{Section~\ref{subsec:comparison}}{the main paper}).
For each evaluation metric, we plot the performance of all other models against the win rate compared to the ShowHowTo model from the user study.
The results can be seen in Figure~\ref{fig:user-study-corr}, we also show the line of best fit and $R^2$ value.
We find a high correlation across all metrics, especially so for Scene Consistency and Task Faithfulness, confirming that our metrics serve as a good proxy for human preference.

\begin{figure*}[t]
    \centering
    \includegraphics[width=\linewidth]{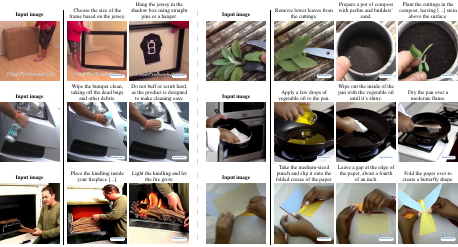}
    \vspace{-0.7cm}
    \caption{\textbf{Qualitative results of our method for sequences from the test set.} Our model can generate both short instructional sequences (as shown here) as well as long or very long sequences shown in Figures~\ref{supmatfig:qualres3} and ~\ref{supmatfig:qualres2}.}
    \label{supmatfig:qualres1}
\end{figure*}

\section{Additional Quantitative Results}\label{supmatsec:addquantres}

\begin{table}[t]
\centering
\scriptsize
\begin{tabular}{lc}
\toprule
Method  & FID$\downarrow$ \\
\midrule
InstructPix2Pix~\cite{brooks2023instructpix2pix} & 37.8 \\
AURORA~\cite{krojer2024aurora} & 23.2 \\
GenHowTo~\cite{soucek2024genhowto} & 28.3 \\
Phung \etal~\cite{phung2024coherent} & 27.8 \\
StackedDiffusion~\cite{menon2024generating} & 34.6 \\
\textbf{ShowHowTo} & \textbf{12.4} \\
\bottomrule
\end{tabular}
\caption{\textbf{Comparison with state-of-the-art using the FID score} on the ShowHowTo test set.
ShowHowTo model significantly outperforms all related methods.}
\label{tab:fid-results}
\end{table}

\begin{table}
\centering
\scriptsize
\begin{tabular}{lc}
\toprule
Method & Acc\textsubscript{ac} \\
\midrule
Stable Diffusion~\cite{rombach2022high} & 0.51 \\
Edit Friendly DDPM~\cite{huberman2024edit} & 0.60 \\
InstructPix2Pix~\cite{brooks2023instructpix2pix} & 0.55 \\
GenHowTo~\cite{soucek2024genhowto} & 0.66 \\
\textbf{ShowHowTo} & \textbf{0.72} \\
\bottomrule
\end{tabular}
\caption{\textbf{Zero-shot evaluation on the GenHowTo dataset} according to the GenHowTo protocol~\cite{soucek2024genhowto}. The ShowHowTo model outperforms the prior state-of-the-art with no fine-tuning on the GenHowTo train set.}
\label{tab:genhowto-benchmark}
\smallskip
\end{table}

\paragraphcustomWOvspace{FID results.}
We also evaluate all methods using the FID score in Table~\ref{tab:fid-results}. For each input image $I_0$ and the textual instructions $\{\tau_i\}_{i=0}^n$ from the ShowHowTo test set, we generate the sequence $\{\hat{I}\}_{i=1}^n$ of visual instructions. For each method, its FID score is computed between its generated sequences and the source visual instruction sequences $\{{I}_i\}_{i=1}^n$ from the test set.
Our method generates images that better match the distribution of real visual instruction sequences.

\paragraphcustom{Zero-shot evaluation on the GenHowTo dataset.}
In GenHowTo~\cite{soucek2024genhowto}, the authors propose to evaluate generative methods in a downstream application, where the method generates images of various classes that are then used to train a simple classifier.
The performance of this classifier is then computed on the real set of images.
We show the results of our method in Table~\ref{tab:genhowto-benchmark}. We evaluate our model as is, \ie, trained on the ShowHowTo dataset without any additional training or fine-tuning. We report the action accuracy metric $Acc_{ac}$, which evaluates whether our generated visual instruction images can be used for downstream application of classifying actions.
Note that this metric is image-based and does not evaluate sequences. Nonetheless, ShowHowTo improves over the previous state-of-the-art from~\cite{soucek2024genhowto} by 6 percentage points.

\begin{table}[t]
\centering
\scriptsize
\begin{tabular}{lccccccc} 
 \toprule
 & \multicolumn{7}{c}{Number of generated frames} \\
 \cmidrule(lr){2-8}
 & 1 & 2 & 3 & 4 & 5 & 6 & 7 \\ 
 \midrule
 Step Faith. & 1.00 & 0.72 & 0.59 & 0.52 & 0.50 & 0.50 & 0.51 \\ 
 Scene Consist. & 0.20 & 0.29 & 0.43 & 0.42 & 0.34 & 0.36 & 0.31 \\ 
 Task Faith. & 0.30 & 0.49 & 0.48 & 0.45 & 0.42 & 0.46 & 0.37 \\ 
 \bottomrule
\end{tabular}
\caption{\textbf{Model's performance for different lengths of the generated sequence.} The performance is fairly similar across different lengths, with a decrease observed for longer sequences.}
\label{tab:variablelentest}
\end{table}

\paragraphcustom{Variable sequence length generation analysis.} \added{Our model generates sequences of variable length. We analyze the performance of the model for various sequence lengths in Table~\ref{tab:variablelentest}. Except for the degenerative case with one frame, the performance is fairly similar across different lengths, with a decrease observed for longer sequences.}

\begin{table}[t]
\centering
\scriptsize
\begin{tabular}{lcccc}
\toprule
\multirow{2}{*}{Task category}   & Step & Scene & Task & \multirow{2}{*}{Average}  \\
& {Faithf.} &  {Consist.} & {Faithf.} & \\
\midrule

Cars \& Other Vehicles         & 0.34 & 0.37 & 0.58 & 0.43 \\
Education and Communications   & 0.37 & 0.55 & 0.24 & 0.39 \\
Food and Entertaining          & 0.67 & 0.24 & 0.40 & 0.44 \\
Health                         & 0.29 & 0.56 & 0.46 & 0.44 \\
Hobbies and Crafts             & 0.39 & 0.46 & 0.41 & 0.42 \\
Holidays and Traditions        & 0.56 & 0.35 & 0.52 & 0.48 \\
Home and Garden                & 0.40 & 0.41 & 0.38 & 0.40 \\
Pets and Animals               & 0.48 & 0.41 & 0.39 & 0.43 \\
Sports and Fitness             & 0.36 & 0.42 & 0.55 & 0.44 \\

\bottomrule
\end{tabular}
\caption{\textbf{Model's performance for different task categories.} Task categories with only a few sequences not shown.}
\label{tab:percategoryresults}
\end{table}

\paragraphcustom{Per-task performance analysis.} \added{We report the performance of our model in various HowTo100M task categories in Table~\ref{tab:percategoryresults}. We observe that the performance is significantly dependent on the task distribution in the dataset. In detail, for the Step Faithfulness metric, the best performance is achieved in cooking tasks because of plentiful training data and clear, visually distinct steps. On the other hand, for Scene Consistency, the cooking tasks perform the worst as the matching is done across the whole dataset, where a large portion is cooking with many similar scenes and frames. Additionally, the cooking tasks contain many close-up scenes without any scene background that can be used for matching the generated images to the correct sequence. The best Scene Consistency is achieved for tasks in the Health and the Education categories due to the uniqueness of the sequences in those categories.}

\begin{figure*}[t]
    \centering
    \includegraphics[width=\linewidth]{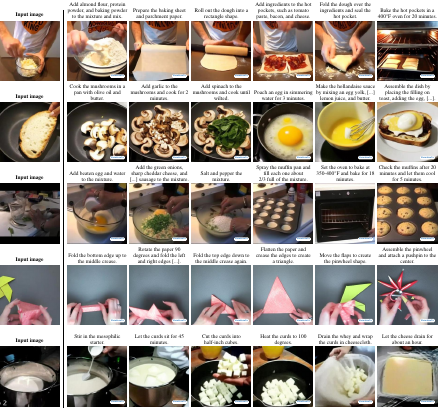}
    \vspace{-0.7cm}
    \caption{\textbf{Additional qualitative results of our method for sequences from the test set.} Given the input image (left) and the textual instructions (top), ShowHowTo generates step-by-step visual instructions while maintaining objects from the input image.}
    \label{supmatfig:qualres3}
\end{figure*}

\begin{figure*}[p]
    \centering
    \includegraphics[width=\linewidth]{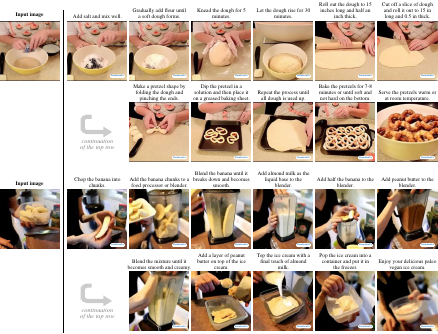}
    \caption{\textbf{Qualitative results of our method for sequences from the test set.} Our model can generate long sequences of visual instructions while being consistent with the input image and the text prompts.}
    \label{supmatfig:qualres2}
\end{figure*}

\begin{figure*}[t]
    \centering
    \includegraphics[width=\linewidth]{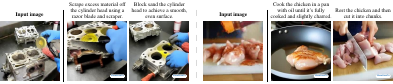}
    \caption{\textbf{Failure modes.} The model can struggle with rare objects and tools (left), or it can fail to update object states after state-changing actions (right).}
    \label{supmatfig:failures}
\end{figure*}

%\vfill\eject

\section{Additional Qualitative Results}\label{supmatsec:qualres}
\paragraphcustomWOvspace{Additional qualitative results.} We show additional qualitative results in Figures~\ref{supmatfig:qualres1}, \ref{supmatfig:qualres3}, and \ref{supmatfig:qualres2}. We show our method can correctly generate sequences of visual instructions according to the input images and prompts. In Figure~\ref{supmatfig:qualres2}, we demonstrate our method can generate long instructional sequences while preserving consistency with the input image. Additionally, in Figure~\ref{supmatfig:qualres1}, we show our model can also generate shorter sequences.

\paragraphcustom{Additional qualitative comparison with related work.}
We show additional comparison with related work on the task of creating paper flowers in Figure~\ref{supmatfig:addqualcomp}. We can see our method not only correctly captures the scene, which is not the case for the method of Phung~\etal~\cite{phung2024coherent} and StackedDiffusion~\cite{menon2024generating}, but the model faithfully follows the input prompts, generating useful visual instructions for the user.

\paragraphcustom{Failure modes.} We show some limitations
of our method, as described in the main paper, in Figure~\ref{supmatfig:failures}.
Our model can struggle with objects that are not common in the training data, such as engine cylinders and tools such as razor blades. Additionally, the model can make errors in scenarios where object states need to be tracked and updated across multiple frames, such as after cooking meat, the meat's state must be changed from \textit{raw} to \textit{cooked}, \etc.

\begin{figure*}[p]
    \centering
    \includegraphics[width=\linewidth]{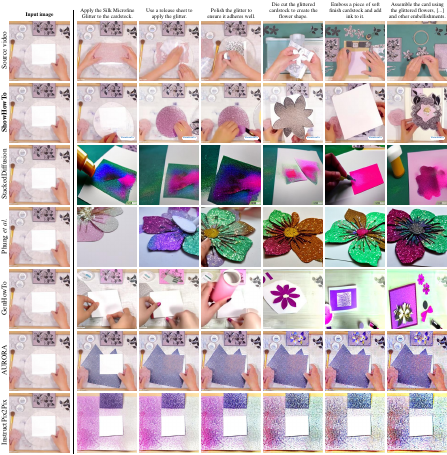}
    \caption{\textbf{Additional qualitative comparison} using the input image (left) and the textual instructions (top) for the task of making a cardboard flower with glitter. Only ShowHowTo can produce convincing steps while preserving the input scene.}
    \label{supmatfig:addqualcomp}
\end{figure*}

\begin{figure*}[p]
    \centering
    \input{supmat_assets/asr_example}
    \caption{Comparison between WhisperX~\cite{bain2021frozen} speech transcription (left) and YouTube ASR (right) for the same \textit{`How to bake a potato in the pressure cooker'} video. In contrast to YouTube ASR, WhisperX can correctly split the narrations into individual sentences. It also makes significantly fewer errors; for example, it correctly recognizes that the potatoes should be cooked for 15 minutes, not 50 minutes (timestamp 59.85).}
    \label{supmatfig:asr_comparison}
\end{figure*}

\begin{figure*}[htbp]
    \centering
    \begin{tcolorbox}[
        colback=white,
        colframe=gray!30,
        coltitle=black,
        title={Prompt for filtering non-instructional videos},
        sharp corners,
        boxrule=0.5pt,
        width=\textwidth,
        enlarge left by=0mm,
        enlarge right by=0mm
        ]

    % \scriptsize
    \footnotesize
    Based on the following video title and partial transcript segment, determine if the video is instructional in nature, where ``instructional'' means it involves actively demonstrating or teaching how to perform a specific task or activity with physical steps (e.g., cooking a recipe, repairing something, crafting, etc.). 
    Respond with `Yes' if the video is actively demonstrating or teaching how to perform a specific task, or `No' if it is not. Then provide a single sentence explanation.\\
    
    Examples of instructional videos: \\
    • How to Bake a Chocolate Cake \\
    • Repairing a Leaky Faucet \\
    • Learn to Knit a Scarf \\
    
    Examples of non-instructional videos: \\
    • Discussing Fashion Trends \\
    • Product Reviews and Opinions \\
    • A Vlog of My Daily Life \\
    
    Example 1: \\
    Video Title: \textit{Red Dead Redemption 2 - Herbert Moon and Strange man Eastereggs In Armadillo [SPOILERS]} \\
    Video Transcript: \textit{``oh you're back I feared the worst it's all here waiting for you who's that I don't know it's just a little portrait somebody gave me once I always quite liked it why no reason just seem familiar anyway this area is closed to the public if you want to shop here you better act right move you long streak of piss who do you think you are for God's sake get out you degenerate you blew it get out of my store if you don't leave there will be problems okay okay stay calm oh you'll (...)''} \\
    Is this video actively demonstrating or teaching how to perform a specific task? No \\
    Explanation: The video is not actively demonstrating or teaching how to perform a specific task; it appears to be showcasing or discussing Easter eggs in the game Red Dead Redemption 2. \\
    
    Example 2: \\
    Video Title: \textit{Fantastic VEGAN Cupcakes with Raspberry Frosting} \\
    Video Transcript:
    \textit{``hey there I'm chef Annie and tomorrow is Valentine's Day so we are making some extra special cupcakes for this occasion can you believe that we have not made cupcakes on this channel it's about time so today I'm going to show you how to present these cupcakes so they look impressive and absolutely beautiful so enough copy let's cook it so we're going to start by mixing together our wet ingredients (...)''} \\
    Is this video actively demonstrating or teaching how to perform a specific task? Yes \\
    Explanation: The video actively demonstrates and teaches how to make vegan cupcakes with raspberry frosting, as indicated by the detailed steps and instructions given by the chef. \\

    Example 3: \\
    Video Title: \textit{How To: Piston Ring Install} \\
    Video Transcript:
    \textit{``hey it's Matt from how to motorcycle repair comm just got done doing a top end on a YZF 250 or yz250 F and I thought I'd do a quick video on how to install a piston ring the easy way now I've done this in the past too but most people will take the ends here and spread it and put it on but you can potentially damage the ring so an easier way to do that is just to take this right here incident in the groove that you need then you bend one up (...)''} \\
    Is this video actively demonstrating or teaching how to perform a specific task? Yes \\
    Explanation: The video is actively demonstrating or teaching how to install a piston ring, which is a specific task. \\

    Example 4: \\
    Video Title: \textit{Best gas weed eater reviews Husqvarna 128DJ with 28cc Cycle Gas Powered String Trimmer} \\
    Video Transcript:
    \textit{``guys i'm shanley today i'm going to tell you about this straight shaft gas-powered trimmer from husqvarna this trimmer runs on a 28 CC two cycle engine it features 1.1 horsepower and a three-piece crankshaft it also has a smart start system as well as an auto return to stop switch and this trimmer is air purge design for easier starting it has a 17 inch cutting path (...)''}\\
    Is this video actively demonstrating or teaching how to perform a specific task? No \\
    Explanation: This video is reviewing the features of a gas-powered trimmer rather than actively demonstrating or teaching how to use it. \\
    
    Now, determine if the following video is instructional in nature:
    
    Video Title: \textcolor{blue}{\{Input video title\}}
    
    Video Transcript:
    \textcolor{blue}{\{Input video transcript\}}
    
    Is this video actively demonstrating or teaching how to perform a specific task?
    \end{tcolorbox}
    \caption{Prompt used for filtering non-instructional videos using Llama 3. Transcript excerpts are truncated for clarity, the full prompt \ARXIVversion{is available on the project website.}{will be released with the code.}}
    \label{supmatfig:prompt-filtering}
\end{figure*}
\begin{figure*}[t]
    \centering
    \begin{tcolorbox}[
        colback=white,
        colframe=gray!30,
        coltitle=black,
        title={Prompt for step extraction from an instructional video},
        sharp corners,
        boxrule=0.5pt,   % border thicknes
        width=\textwidth,
        enlarge left by=0mm,
        enlarge right by=0mm
        ]

    \footnotesize
    Below are transcripts from YouTube instructional videos and their corresponding extracted steps in a clear, third-person, step-by-step format like WikiHow. Each step is concise, actionable, and temporally ordered as they occur in the video. The steps include start and end timestamps indicating when the steps are carried out in the video. Follow this format to extract and summarize the key steps from the provided transcript. \\

    Example 1:\\
    
    YouTube Video Title: ``\textit{BÁNH TÁO MINI - How To Make Apple Turnovers  | Episode 11 | Taste From Home}'' \\ 
    
    YouTube Video Transcript: \\
    00.87 - 07.79: ``\textit{Hey little muffins, today we will make together a super easy, quick and delicious apple turnovers.}" \\
    07.79 - 09.35: ``\textit{40 minutes for all the process.}" \\
    09.35 - 11.95: ``\textit{Seriously, can someone deny them?}" \\
    11.95 - 13.63: ``\textit{Ok, let's begin.}" \\
    13.63 - 18.82: ``\textit{First of all, combine the apple cubes, lemon juice, cinnamon and sugar in a bowl.}" \\
    26.69 - 29.59: ``\textit{Mixing, mixing, mixing.}" \\
    29.59 - 32.62: ``\textit{Apple and cinnamon always go perfectly together.}" \\
    32.62 - 43.52: ``\textit{Now using a round cutter or glass like me, cut 15 rounds from the pastry sheet.}" \\
    57.86 - 64.99: ``\textit{Here comes the fun part.}" \\
    64.99 - 69.97: ``\textit{Spoon about 2 teaspoons apple mixture in the center of one round.}" \\
    69.97 - 74.41: ``\textit{Using your fingers, gently fold the pastry over to enclose filling.}" \\
    88.47 - 104.48: ``\textit{After that, use a fork and press around the edges to seal and make your apple turnovers look more beautiful.}" \\
    104.48 - 105.84: ``\textit{This is how it looks like.}" \\
    109.99 - 113.53: ``\textit{I will show you one more time to make sure that you understand the technique.}" \\
    113.53 - 117.20: ``\textit{And if you still find my apple turnovers too ugly, I'm really sorry.}" \\
    %117.20 - 121.66: ``Anyways, just have fun making them with your family and friends." \\
    (...)\\
    
Extracted Steps: 
\begin{lstlisting}
[{ "WikiHow Title": "How to Make Apple Turnovers" },
  { "steps": [
    { "step": 1, "instruction": "Combine apple cubes, lemon juice, cinnamon, and sugar in a bowl.", "start_timestamp": 13.63, "end_timestamp": 18.82 },
    { "step": 2, "instruction": "Mix the ingredients thoroughly.", "start_timestamp": 26.69, "end_timestamp": 29.59 },
    { "step": 3, "instruction": "Cut 15 rounds from the pastry sheet using a round cutter or a glass.", "start_timestamp": 32.62, "end_timestamp": 43.52 },
    { "step": 4, "instruction": "Spoon about 2 teaspoons of the apple mixture into the center of one round.", "start_timestamp": 64.99, "end_timestamp": 69.97 },
    { "step": 5, "instruction": "Gently fold the pastry over to enclose the filling using your fingers.", "start_timestamp": 69.97, "end_timestamp": 74.41 },
    { "step": 6, "instruction": "Press around the edges with a fork to seal and beautify the turnovers.", "start_timestamp": 88.47, "end_timestamp": 104.48 },
    { "step": 7, "instruction": "Repeat the technique until all turnovers are formed.", "start_timestamp": 109.99, "end_timestamp": 113.53 },
    { "step": 8, "instruction": "Lightly beat one egg in a small bowl.", "start_timestamp": 151.62, "end_timestamp": 157.46 },
    { "step": 9, "instruction": "Egg wash the apple turnovers to give them a gorgeous light brown color after baking.", "start_timestamp": 157.46, "end_timestamp": 164.10 },
    { "step": 10, "instruction": "Bake the apple turnovers at 180°C for 18-20 minutes until golden.", "start_timestamp": 164.10, "end_timestamp": 174.87 },
    { "step": 11, "instruction": "Enjoy the freshly baked apple turnovers.", "start_timestamp": 178.17, "end_timestamp": 185.55 }]}]
\end{lstlisting}

    Example 2: (...) \\

    Now, extract the steps from the following transcript:\\

    YouTube Video Title:
    \textcolor{blue}{\{Input video title\}}\\
    
    YouTube Video Transcript:
    \textcolor{blue}{\{Input video transcript\}}\\
    
    Extracted Steps:

    \end{tcolorbox}
    \vspace{-0.35cm}
    \caption{Prompt used for generating instructional steps with start-end timestamps using Llama 3. The prompt is truncated for clarity, the full prompt \ARXIVversion{is available on the project website.}{will be released with the code.}}
    \label{supmatfig:prompt-step-extraction}
\end{figure*}
}{}

\end{document}